%% file: main.tex
\documentclass{article}

\usepackage[preprint,nonatbib]{neurips_data_2022}

\usepackage[utf8]{inputenc} 
\usepackage[T1]{fontenc}    
\usepackage{hyperref}       
\usepackage{url}            
\usepackage{booktabs}       
\usepackage{amsfonts}       
\usepackage{nicefrac}       
\usepackage{microtype}      
\usepackage{xcolor}         
\usepackage{xspace}

\newcommand{\ours}{B-FHTL\xspace}
\usepackage{graphicx}
\usepackage{subfig}
\usepackage{threeparttable}
\newcommand{\paratitle}[1]{\vspace{0ex}\noindent\textbf{#1}}
\usepackage{amsthm}
\usepackage{multirow}
\theoremstyle{definition}
\usepackage{wrapfig}

\title{A Benchmark for Federated Hetero-Task Learning}
\author{%
  Liuyi Yao, Dawei Gao, Zhen Wang,  
  Yuexiang Xie, Weirui Kuang,
  Daoyuan Chen, \\
  \textbf{Haohui Wang, Chenhe Dong, Bolin Ding, Yaliang Li\thanks{Corresponding author}}\\
  Alibaba Group\\
  \texttt{\{yly287738, gaodawei.gdw, jones.wz, yuexiang.xyx\}@alibaba-inc.com}\\
  \texttt{\{weirui.kwr, daoyuanchen.cdy, wanghaohui.whh\}@alibaba-inc.com}\\
  \texttt{\{dongchenhe.dch, bolin.ding, yaliang.li\}@alibaba-inc.com}
}

\begin{document}

\maketitle
\begin{abstract}

To investigate the heterogeneity in federated learning in real-world scenarios, we generalize the classic federated learning to federated hetero-task learning, which emphasizes the inconsistency across the participants in federated learning in terms of both data distribution and learning tasks. We also present \textbf{\ours}, a federated hetero-task learning benchmark consisting of simulation dataset, FL protocols and a unified evaluation mechanism. \ours dataset contains three well-designed federated learning tasks with increasing heterogeneity. 
Each task simulates the clients with different non-IID data and learning tasks. To ensure fair comparison among different FL algorithms, \ours builds in a full suite of FL protocols by providing high-level APIs to avoid privacy leakage, and presets most common evaluation metrics spanning across different learning tasks, such as regression, classification, text generation and etc. Furthermore, we compare the FL algorithms in fields of federated multi-task learning, federated personalization and federated meta learning within \ours, and highlight the influence of heterogeneity and difficulties of federated hetero-task learning. Our benchmark, including the federated dataset, protocols, the evaluation mechanism and the preliminary experiment, is open-sourced at
\hyperlink{https://github.com/alibaba/FederatedScope/tree/master/benchmark/B-FHTL}{https://github.com/alibaba/FederatedScope/tree/master/benchmark/B-FHTL}.

\end{abstract}

\input{subfile/1_Introduction}

\input{subfile/2_Federated_Heterogeneous_TL}

\input{subfile/3_Benchmarks}

\input{subfile/4_Experiment}

\input{subfile/5_Conclusion}

\bibliographystyle{abbrv}
\bibliography{benchmark}
\newpage
\appendix
\input{subfile/appendix}

\end{document}

%% file: subfile/1_Introduction.tex
\section{Introduction}

Nowadays, with the increasing public concern over the privacy leakage in machine learning, Federated Learning (FL), which collaboratively trains the machine learning model without directly sharing the raw data among the data holders, has become a trending solution for privacy-preserving computation in both academia and industry~\cite{yang2019federated_app,hard2018federated,xu2021federated, leroy2019federated}. Various federated algorithms, platforms, and benchmarks have emerged aiming to promote real-world federated learning applications~\cite{yang2019federated,pysyft,fedml,tff,federated_scope,leaf}. 
\begin{figure}[ht]
\centering
\subfloat [Classic FL.]{
\includegraphics[width=0.48\textwidth]{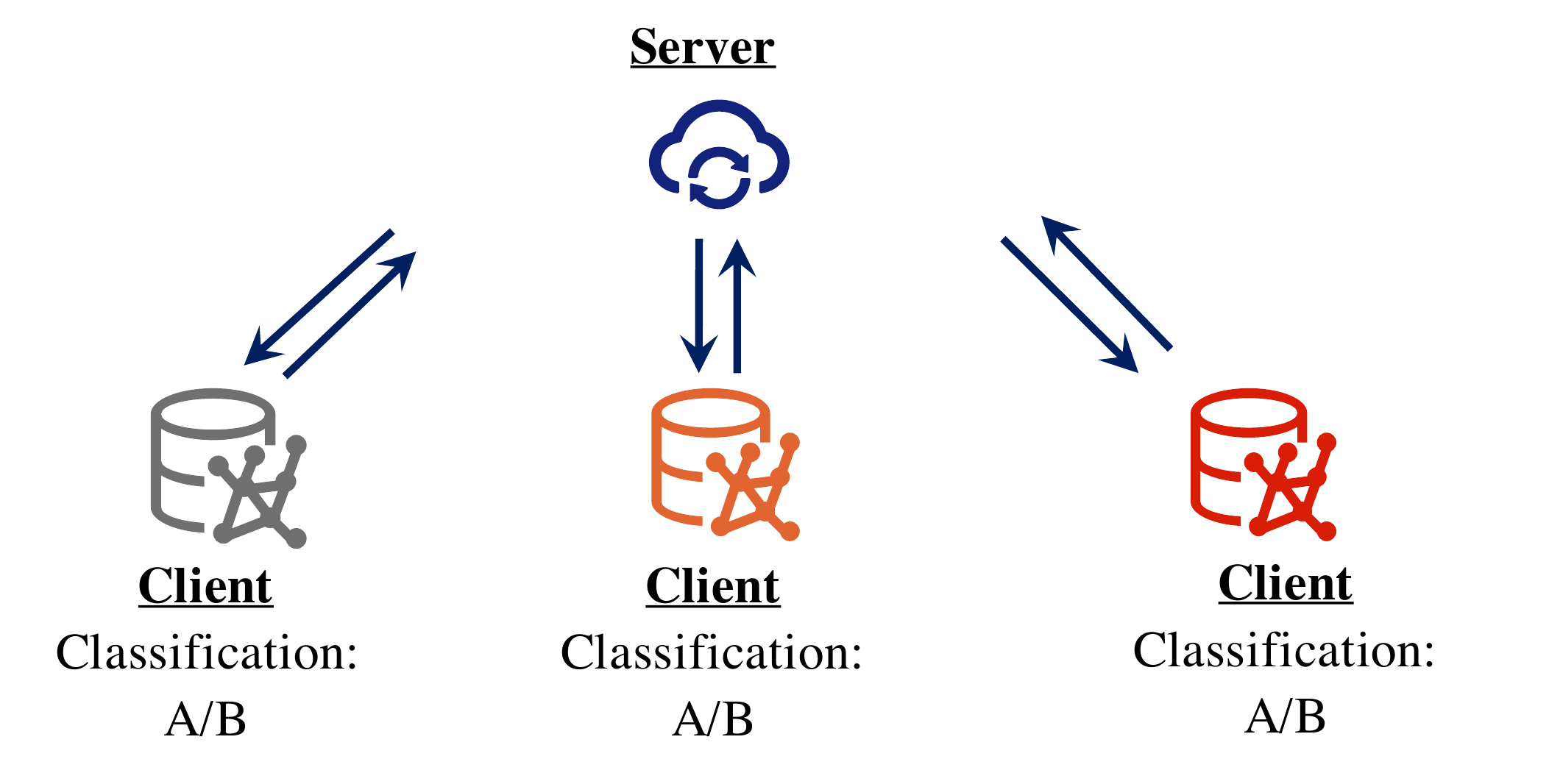}
\label{fig: classic_fl}
}
\hfill
\subfloat [Drug discovery: Heterogeneous class labels.]{
\includegraphics[width=0.48\textwidth]{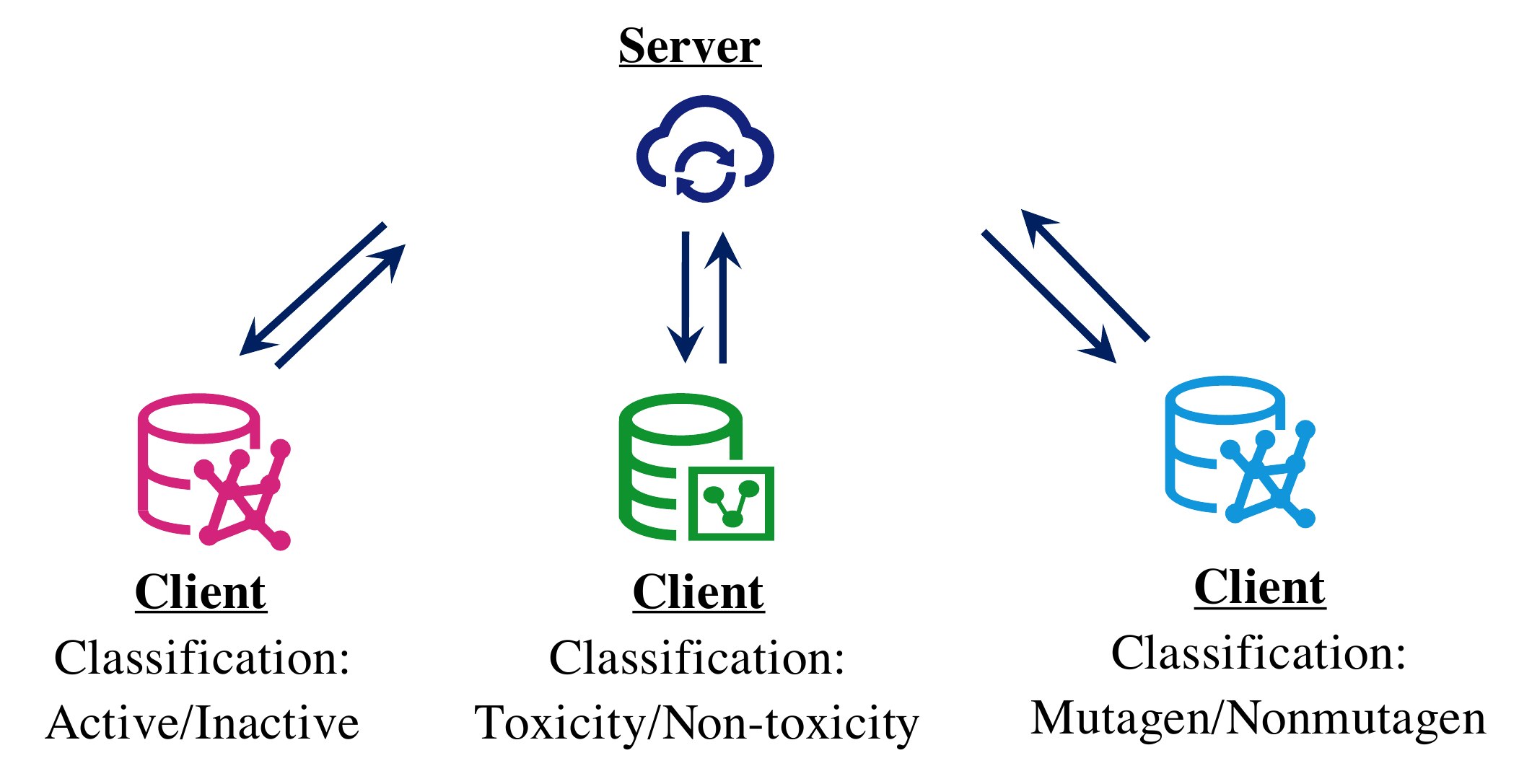}
\label{fig: hetero-class-graph}
}
\hfill
\subfloat [Drug discovery: Heterogeneous task types.]{
\includegraphics[width=0.48\textwidth]{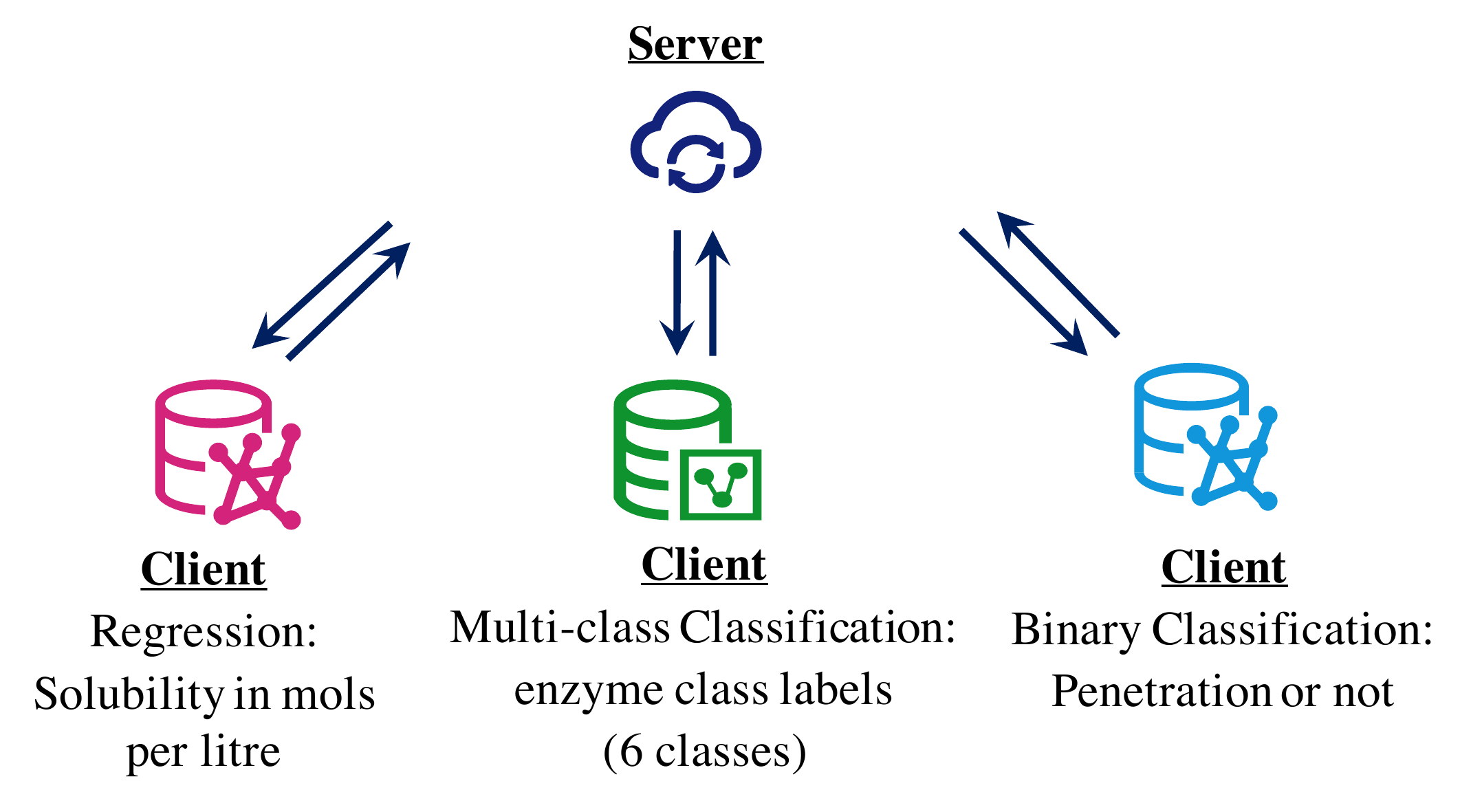}
\label{fig: hetero-task-graph}
}
\hfill
\subfloat [NLP: Heterogeneous task types.]{
\includegraphics[width=0.485\textwidth]{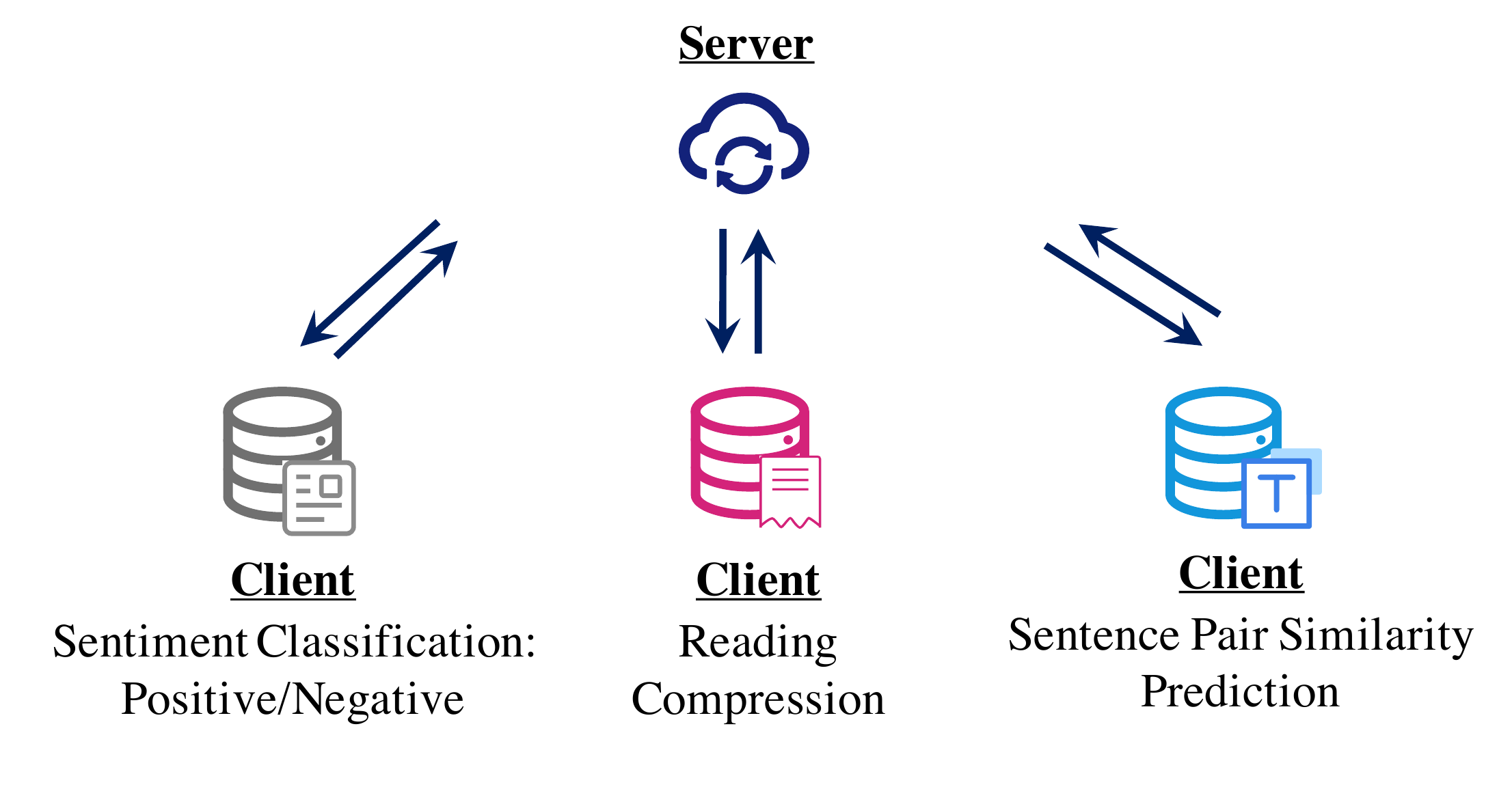}
\label{fig: hetero-task-nlp}
}
\caption{Comparison between classic FL and federated hetero-task learning.}
\label{fig: fhtl}
\end{figure}

One of the key challenges in federated learning is to handle the heterogeneity across different clients. Recently, a great of efforts have been devoted to the problem of data non-IID~\cite{zhao2018federated,zhu2021federated}. However, in real-world applications, the learning goals of different clients are usually different due to various driven businesses. For example, as shown in figure~\ref{fig: hetero-class-graph} institutions join up to explore the chemical properties of different molecular structures, but may diverge on the learning goals due to different business purposes. Although classical FL has been aware of non-IID data among clients, it is still invalid in this scenario due to the assumption that all clients share the same learning tasks. To fulfill the scenarios where clients' learning objectives are heterogeneous, we generalize the classical FL to \emph{federated hetero-task learning}, which takes both data and task heterogeneity into consideration. The goal of the proposed heterogeneous federated learning is to boost the performance across all clients’ learning tasks, since each client contributes to the federal process and is eager to reap benefits from the collaboratively learned model. The new setting is more practical and has great potential to broaden the applications of federated learning since it encourages more institutions with different learning goals to join and benefit from federated learning. Figure~\ref{fig: fhtl} shows the different real-world hetero-task learning applications: Figure~\ref{fig: classic_fl} is the classic federated learning with non-IID data under the same learning objective; Figure~\ref{fig: hetero-class-graph} is a representative application of federated hetero-task learning in the drug discovery area, where institutions with different moleculenets have heterogeneous classification labels; Figure~\ref{fig: hetero-task-graph} shows an extension of the previous example where the task types are heterogeneous including both classification and regression task; Figure~\ref{fig: hetero-task-nlp} shows the applications in natural language processing (NLP) area, where clients with different NLP tasks such as sentiment classification, reading compression, and sentence pair similarity prediction, collaboratively join the federated learning to gain the benefit for their own tasks. Furthermore, federated hetero-task learning brings more real-world applications to federated learning and enlarges the scope of federated learning. Specifically, the heterogeneous task setting enables the domains, which were previously unsuitable for classic FL with the homogeneous task, to develop in the federated format, such as model pre-training~\cite{devlin2018bert,dong2019unified}, life-long learning~\cite{parisi2019continual}, web3.0~\cite{hendler2009web} and multimodal learning~\cite{ramachandram2017deep}.

However, there are three challenges that obstacles the fast development of federated hetero-task learning. (1) The first one is the lack of federated datasets that imitate real-world federated hetero-task learning. Existing FL benchmarks either only provide the dataset with the same learning objective or create the heterogeneous tasks by letting different clients hold data belonging to different classes~\cite{leaf}, which only covers parts of real-world federated hetero-task learning. As shown in figure~\ref{fig: fhtl}, the tasks across different clients are different, and they are not limited to the same type of task. Existing benchmark datasets fail to represent such a real-world hetero-task setting. (2) The second challenge is the lack of federated learning protocols to ensure the developed method follows the FL privacy requirements. Some trivial solutions may sacrifice privacy for performance improvement; for example, the private data of different clients may directly be fetched and aggregated. Therefore, without a unified FL protocol, it is hard to determine whether the developed method is suitable for the federated hetero-task learning and conduct a fair comparison between different methods. (3) The third challenge is the lack of an evaluation mechanism for a fair comparison. Due to the fact that in hetero-task learning, the task type can be different, it is highly possible that different clients have different evaluation metrics. How to aggregate the performance results among all clients for fair comparison still remains an open question.

To overcome the above challenges and promote the development of federated hetero-task learning, we propose \textbf{\ours}, a benchmark framework for federated hetero-task learning. 
Our benchmark provides a comprehensive simulation for federated hetero-task learning, which embraces datasets, protocols, and the evaluation mechanism.
Specifically, \ours contains three well-designed federated datasets collected from graph and NLP domains containing different tasks, such as graph classification, regression, sentiment classification, reading compression and sentence pair similarity prediction.
Furthermore, \ours builds a series of FL protocols based on FederatedScope~\cite{federatedscope} to regularize the shared information among different clients, protect the data from leakage and promise a fair comparison among various FL methods. 
\ours also provides high-level APIs to simplify the FL deployment and shield developers from implementation details. In terms of evaluation mechanism, since evaluation metrics for federated hetero-task learning are still in the early stage, \ours integrates various types of evaluation metrics for comprehensive fair comparisons, such as weighted performance average, and also supports client-level observation.

To the best of our knowledge, this is the first federated hetero-task learning benchmark that provides the all-around simulation for convenient FL methods implementation and fair comparison. Our benchmark can encourage the interdisciplinary research of federated learning with Multi-task Learning, Model Pre-training, and AutoML (including Meta-Learning and Hyperparameter Optimization) as they are closely related to federated hetero-task learning. Meanwhiles, our benchmark's setting is a more practical setting to enlarge the real-world application scope of Federated Learning.
Overall, the contributions of our proposed benchmark is: 
\begin{itemize}
    \item We generalize the classical FL setting to the federated hetero-task learning setting, which takes both the data heterogeneity and learning goal heterogeneity into consideration. This setting is more practical and can promote a wide range of real-world federated learning applications.
    \item We propose the first federated hetero-task learning benchmark, including the federated datasets, the protocols, and the evaluation mechanism, which comprehensively simulates the real-world FL applications. The provided benchmark is easy-to-use and ensures fair comparison among different methods.
    \item We present the experiment analysis including four types of methods and highlight the potential of interdisciplinary research of Federated Learning with Multi-task Learning, Model Pre-training, and AutoML in promoting the development of federated hetero-task learning. 
\end{itemize}

%% file: subfile/2_Federated_Heterogeneous_TL.tex
\section{Related Work}

\paratitle{FL Benchmark.} With the increasing demand for federated learning, there are several existing federated learning benchmarks for different purposes. To change the status quo of the lack of consistent comparison mechanisms, the benchmark FedEval~\cite{chai2020fedeval} provides an evaluation platform, which enables the comparison under the same condition. To facilitate the development of federated learning with clients heterogeneous in data distribution, data scale, and device connection, FedScale~\cite{lai2021fedscale} is proposed to enable efficient FL evaluation. Furthermore, FedGraphNN~\cite{FedGraphNN} is a benchmark proposed for federated graph learning and in~\cite{clinic_fl_benchmark}, a benchmark for federated learning on clinical data is provided to examine the existing FL works in the clinic domain. Last but not least, LEAF~\cite{leaf} which contains a series of federated datasets, an evaluation framework and a set of baseline implementations, is the most similar benchmark to ours. However, compared with above benchmarks, \ours focuses on the scenario where clients' learning objectives are heterogeneous, and provides three different federated datasets reflecting different heterogeneous levels. Further, \ours also provides a high-level API embedding FL protocols and evaluation metrics ensuring reproducibility and fair comparison.

\paratitle{Federated Hetero-Task Learning.}  
The most relevant existing work to the federated hetero-task learning is the federated multi-task learning, which considers data heterogeneity and task heterogeneity. To solve these challenges, based on the multi-task learning framework, a system-aware optimization method MOCHA~\cite{FMTL} is proposed, which not only models the task relationships among different clients, but also is able to tolerate the computation and communication heterogeneity. HeteroFL~\cite{heteroFL} further solves the system challenge by allowing the clients collaboratively train the heterogeneous local models that are much smaller than the global model. Personalized federated learning is also a related topic, whose objective is to improve performance in the case where clients' data are non-IID. Existing personalized federated learning works solve the data heterogeneity challenge by model mixture \cite{zhang2020personalized,liDittoFairRobust2021}, clustering \cite{briggs2020federated,sattler2020clustered}, knowledge distillation \cite{linEnsembleDistillationRobust2021,zhuDataFreeKnowledgeDistillation2021}, meta-learning \cite{khodak2019adaptive,jiangImprovingFederatedLearning2019,khodak2019adaptive,fallahPersonalizedFederatedLearning2020}, and transfer learning \cite{yangFedStegFederatedTransfer2020,annavaramGroupKnowledgeTransfer,zhang2021parameterized}. However, as personalized federated learning assumes all clients have the same learning objective, the existing personalized federated learning works is limited in the federated hetero-task learning setting.

\section{Problem Definition}

The target of federated hetero-task learning is to learn some common knowledge cooperatively without the consistency assumptions in both data  and task. Similar with the general federated learning, we name the participants holding the data as \emph{client}, and the participant coordinating the clients as \emph{server}.
In federated hetero-task learning each client samples data from different distribution and holds different learning goals.
To be specific, the problem is defined as follows:
\begin{itemize}
    \item \textbf{Input}: Several clients, each one is associated with a different dataset (feature space is aligned) and a different learning objective.
    \item \textbf{Output}: A learned model for each client, and a central model across clients (this central model is the outputted model in traditional federated learning).
    \item \textbf{Objective}: Performance improvements across all clients’ learning tasks.
\end{itemize}

%% file: subfile/3_Benchmarks.tex
\section{Benchmark for Federated Hetero-Task Learning}
\begin{wrapfigure}[13]{r}{0.5\textwidth}
\centering
\includegraphics[width=0.5\textwidth]{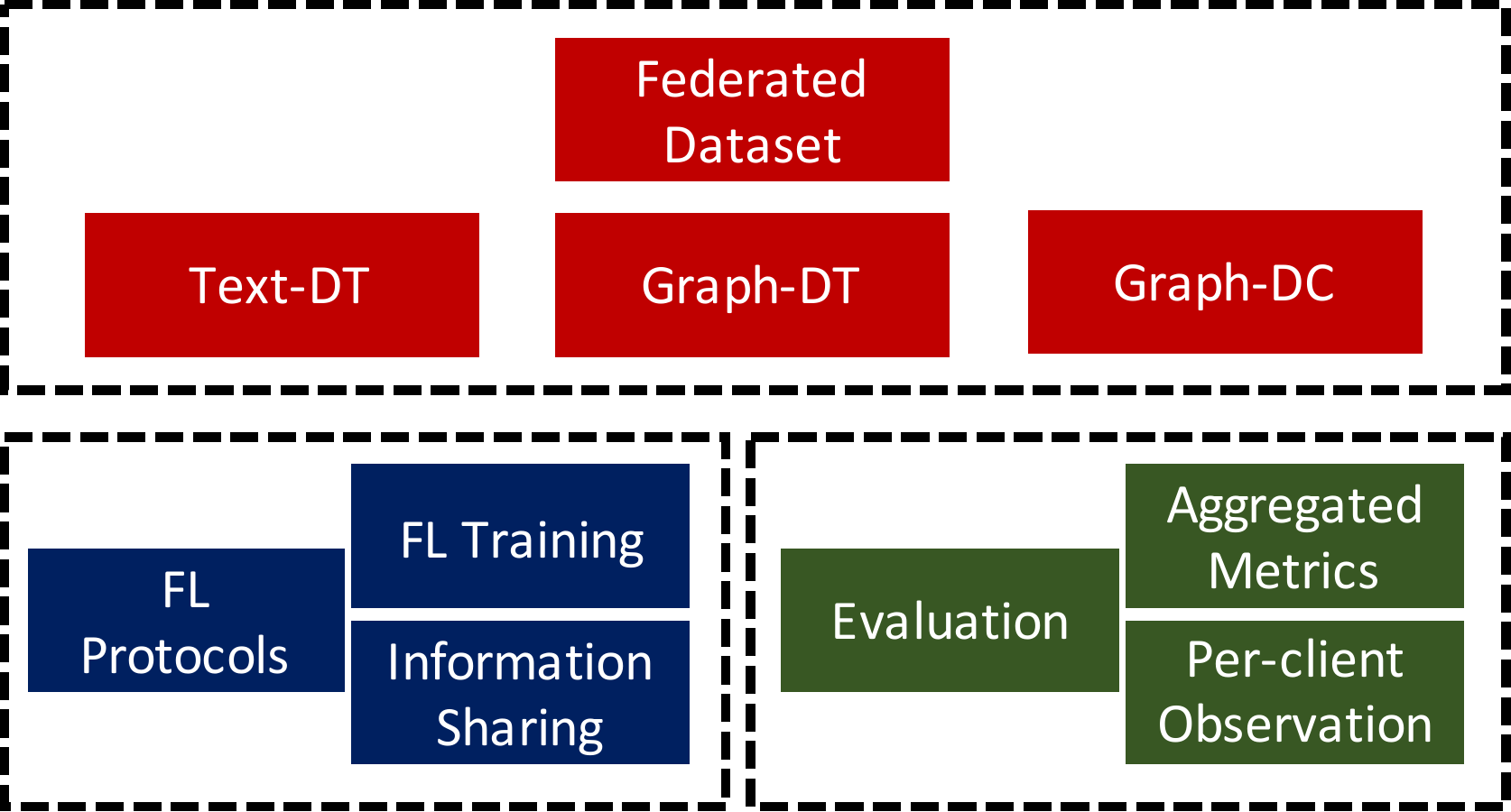}
\caption{Overview of \ours.}
\label{fig:overview}
\end{wrapfigure}\textbf{Overview.} 
Figure~\ref{fig:overview} shows the major components of \ours. In federated dataset components, we introduce three datasets: (1) Graph-DC comes from the graph domain, and in this dataset, different clients hold the data belonging to a different class. In this dataset, all clients still have the same task type graph classification, and it is quiet common in the existing FL benchmarks~\cite{leaf,lai2021fedscale} that explore the federated multi-task learning; (2) Graph-DT is also from the graph domain. Compared with Graph-DC, clients in Graph-DT hold tasks with different types: some clients hold classification tasks, and some hold regression tasks. (3) Text-DT has the same client heterogeneity setting as Graph-DT, and comes from NLP domain. In FL protocols components, we ensure the development of innovative methods to follow the privacy-preserving requirements by regularizing the federated training and the information sharing among clients/server. In the evaluation component, \ours provides various evaluation metrics, including the aggregated metrics and the client-level metrics. In the following, we introduce the three components in detail.

\subsection{Data Curation}
\label{sec:dataset}

\paratitle{Graph Dataset with Distinct Classes.} The first dataset simulates the case shown in figure~\ref{fig: hetero-class-graph} where the tasks across different clients are heterogeneous in terms of class labels. For simplicity, we name this dataset as \textbf{Graph-DC}, which is shorted for \textbf{g}raph dataset with \textbf{d}istinct \textbf{c}lasses. In this dataset, there are total $13$ clients, and each client privately holds a graph classification data with distinct binary classes. Specifically, each client's data comes from one specific public graph classification dataset, which imitates the federated hetero-task learning setting where clients with different learning goals collect their data independently. Among the 13 clients, $11$ clients are selected from TUDataset~\cite{Morris+2020}, a collection of graph classification dataset collection, and the remaining two clients are created by separately sampling $1000$ records from HIV and BACE dataset in MoleculeNet~\cite{moluculenet}, which is a benchmark for molecular machine learning.
Table~\ref{tab:client_statistics} lists the statistics of each client, where the clients are numbered by the ascending order of total number of graphs they own. 
\begin{table}[ht]
    \centering
    \begin{tabular}{c c c c cc c}
    \toprule
    Client \# & Dataset Name & Task Type  &  No. Graphs  & Average $|V|$  & Average $|E|$  & \\
    \midrule
    1& MUTAG  & Binary Classification& $188$  &  $17.93$ &  $39.58$  & \\
    2&PTC\_MM  & Binary Classification &$336$  &  $13.97$  &  $28.64$  & \\
    3 & PTC\_MR  &Binary Classification  &$344$  &  $14.29$  &  $29.38$  & \\
    4&PTC\_FR  & Binary Classification &$351$  &  $14.56$  &  $30.01$  & \\
    5& BZR  & Binary Classification  &$405$  &  $35.75$  &  $76.72$  & \\
    6& COX2  & Binary Classification &$467$ &  $41.22$  &  $86.89$   & \\
    7& DHFR  & Binary Classification &$756$ &  $42.43$  &  $89.09$  & \\
    8&HIV  & Binary Classification &$1000$  &  $25.74$  &  $55.45$  & \\
    9&BACE  & Binary Classification &$1000$  &  $33.93$  &  $73.36$  & \\
    10&AIDS  &Binary Classification  &$2000$ &  $15.69$  &  $32.39$ & \\
    11&NCI1  & Binary Classification &$4110$ &  $29.87$  &  $64.60$  & \\
    12& NCI109  &Binary Classification  &$4127$  &$29.69$  &  $64.26$  & \\
    13&Mutagenicity  & Binary Classification &$4337$  &  $30.32$  &  $61.54$  & \\
    \midrule
  Total& -  &  &$19421$ &  $28.41$ &  $60.14$ &\\
    \bottomrule
    \end{tabular}
    \caption{The statistics of dataset Graph-DC.}
    \label{tab:client_statistics}
\end{table}

\paratitle{Graph Dataset with Different Task Types.} The second dataset simulates the case where tasks across different clients are different in terms of task types, as shown in figure~\ref{fig: hetero-task-graph}. We name this dataset as \textbf{Graph-DT}. Specifically, in the Graph-DT dataset,
there are total $16$ clients, with $10$ clients holding the binary classification task, and $6$ clients owning the regression task. Each client's data are from one specific graph dataset selected from graph benchmarks TUDataset and MoleculeNet. 
The details of those tasks are shown in Table~\ref{tab: GDT}. In the table, the first $10$ clients hold the classification task, and they are numbered in ascending order of data size. The remaining clients hold the regression task, and they are numbered in the same way as the first $10$ clients.
\begin{table}[ht]
    \centering
    \begin{tabular}{c c c c cc c}
    \toprule
    Client \# & Name. Dataset & Learning goal  &  No. graphs  & Average $|V|$  & Average $|E|$  & \\
    \midrule
1 & MUTAG   & Binary Classification &188    &17.93& 19.79\\
2 & PTC\_MM & Binary Classification &336    &13.97& 14.32\\
3 & PTC\_MR & Binary Classification &344    &14.29& 14.69\\
4 & PTC\_FM    & Binary Classification &349    &14.11& 14.48\\
5 & PTC\_FR    & Binary Classification &351    &14.56& 15.00\\
6 & ClinTox & Binary Classification &1478   &21.16& 55.76\\
7 & BACE    & Binary Classification &1513   &34.09& 73.72\\
8 & BBBP    & Binary Classification &2039   &24.06& 51.91\\
9 & NCI1   & Binary Classification &4110   &29.87  &32.30\\
10 & NCI109 & Binary Classification &4127   &29.68& 32.13\\
11 & FreeSolv    & Regression    &642    &8.72   &16.78\\
12 & ESOL    & Regression    &1128   &13.29& 27.35\\
13 & Lipophilicity   & Regression    &4200   &27.04  &59\\
14 & QM9    & Regression with 19 values     &129433 &18.03  &18.63\\
15 & alchemy\_full  & Regression with 12 values &202579 &10.10  &10.44\\
16 & ZINC\_full & Regression    &249456 &23.14  &24.91\\
\bottomrule
    \end{tabular}
    \caption{The statistics of dataset Graph-DT.}
    \label{tab: GDT}
\end{table}

\begin{table}[ht]
    \centering
    \begin{tabular}{cccc}
    \toprule
     Client \#     & Name. Dataset & Learning goal & No. records \\
     \midrule
    1 & STS-B~\cite{cer-etal-2017-semeval} & Sentence pair similarity prediction & $7140$\\
    2    &  IMDB~\cite{IMDB_dataset} & Sentiment classification & $50000$\\
    3 & SQuAD (v2.0)~\cite{SQuADv2} & Reading compression & $151054$\\
     \bottomrule
    \end{tabular}
    \caption{The statistics of dataset Text-DT}
    \label{tab: text-DT}
\end{table}
\paratitle{Text Dataset with Different Task Types.} The third dataset comes from NLP domain. In this dataset, there are total $3$ clients, with each holding sentiment classification, reading compression ( finding the answer span in the paragraph given a question), and sentence pair similarity prediction, separately. We name this dataset as \textbf{Text-DT}. Table~\ref{tab: text-DT} summarizes the statistics of each client's data. Specifically, the data of sentiment classification comes from IMDB Review dataset~\cite{IMDB_dataset}, containing $50000$ movie reviews from IMDB; The data of reading compression task comes from SQuAD (v2.0) dataset~\cite{SQuADv2}, containing 151054 records collected from Wikipedia; The data of sentence pair similarity prediction comes from STS benchmark~\cite{dataset_abstractive_sum}, where each pair is annotated with a similarity score from 1 to 5. Compared with Graph-DT, the tasks in the text dataset are not limited to classification/regression.

\subsection{Protocols}
In federated learning, protecting private data is necessary when developing new methods. To ensure the implemented methods follow the privacy-preserving requirements, we embrace the necessary protocols into a federated learning framework to regularize the private data accessibility, the federated learning method training, and the information shared among the clients and the server.

\paratitle{FL Training.} 
Based on FederatedScope~\cite{federated_scope}, \ours provides high-level APIs for users to implement their codes that customize how to train models on these training data without data leakage. After implementation, the FederatedScope will simulate the federated learning, including setting up a server and several clients, local training, communication among clients and the server, and information aggregation. The users can request necessary training statistics, including the training/valid loss, training/valid accuracy, and other customized information. However, it is forbidden to request the highly private-sensitive data (e.g., the raw data) to be shared away from the data owner.

\paratitle{Information Sharing} 
The above protocol regularizes the training environment and data accessibility of the users. In federated learning, the information shared among the clients and the server should also follow the privacy-preserving requirement when developing methods. To fulfill such requirement, a white list of information types that can be shared among clients and the server in federated training is formally defined and set up in our developed benchmark, which prevents sending the private training data directly. Specifically, the information allowed to be shared includes: (1) The data statistics, such as the average, median, count, etc; (2) The model parameters; (3) The aggregation weight in FL process; (4) The gradients (including high-order gradients). The above-allowed information types cover the necessary elements required to implement the existing federated learning algorithms, and ensure that the compared methods are under the FL scenario.

The backend of \ours, FedearetdScope framework, supports the white-list type information sharing protocol due to its built in communications module. In our provided API, the users only need to change the \texttt{trainer} module to develop their own methods, which indicates that when we pre-define the transmission actions, users will naturally follow this protocol. Furthermore, the information types during the transmission will be logged to users, so that the transmission actions are transparent to users. Additionally, to prevent the case of sharing the private data masquerades as allowed-type information, \ours also adopt an communication cost monitor to detect abnormal transmission.

\subsection{Evaluation}
The trained models will be automatically evaluated on each client’s test data independently. As the development of evaluation metrics for federated hetero-task learning is in the early stage, we provide comprehensive evaluation metrics for different heterogeneity settings. 

On Graph-DC dataset, as the tasks of different clients have the same type (binary classification), we provide various ways to aggregate the per-client performance: (1) Equal-weight aggregation: The evaluated results, such as accuracy, are aggregated with equal weights to generate the final evaluation for performance comparison. (2) Data-size related weighted aggregation: the clients' performance results are aggregated by the weights that are proportional to their data size. (3) Customized-weight aggregation: we also support the users to define each client's weights according to their reality needs.

On Graph-DT and NLP-DT datasets, as different task types are associated with different evaluation metrics, it is not very meaningful to directly aggregate them.
Instead, we set up one baseline method, and aggregate the per-client improvement ratio over this baseline. Using an improvements ratio ensures the values to be aggregated are with the same meaning, and the aggregated value denotes the averaged improvements over that baseline. We set the default compared baseline as the "isolated" method where clients only use their own data to generate the model. In this case, the aggregated improvement ratio represents the gain obtained from federated learning. Formally, the overall performance is calculated as:
\begin{equation}
    Overall = I_i \frac{1}{n} \sum_{i=1}^{n} \left(\frac{m_i - b_i}{b_i} \times 100\% \right),
\end{equation}
where $m_i$, $b_i$ is the performance of the developed method and baseline on client $i$, respectively; $n$ is the total number of clients; $I_i$ is the comparison indicator of client $i$, and $I_i = -1$ if the lower evaluation metric value on client $i$ is, such as mean square error (MSE), the better performance; Otherwise $I_i = 1$. In the previous case, when the evaluation metric is mse, $\frac{m_i - b_i}{b_i}$ is usually a negative value when the developed method is better than the baseline method (i.e., $m_i < b_i$). Setting $I_i=-1$ in this case ensures the improvements is a positive value.

Apart from the above aggregated evaluation metrics, \ours also supports the client-level performance observation for all datasets. With the per-client observation, users can clearly get the per-client improvement in detail.

%% file: subfile/4_Experiment.tex
\section{Preliminary Experimental Analysis}
\label{sec:methods}

\textbf{Comparing Methods.} Since federated hetero-task learning is closely relevant to personalized Federated Learning, meta Learning, and multi-task Learning, we extend the state-of-art methods in these fields to our setting. Overall, we primarily use four categories of methods: 
\begin{itemize}
    \item The first category is "isolated", which means that each client produces a model using only its own data. 
    \item The second category includes standard federated optimization algorithms such as FedAvg~\cite{fedavg}, and FedProx~\cite{blocal}, in which clients share only the graph neural network (GNN)~\cite{gin} but have their own classifiers due to the heterogeneity.
    \item Third, personalized federated learning algorithms are particularly well-suited to dealing with heterogeneity and should be included, of which we have tried FedBN~\cite{fedbn} and Ditto~\cite{li2021ditto}.
    \item The fourth category is federated meta-learning, which uses optimization-based meta-learning algorithms rather than standard gradient descent algorithms to perform local updates on the GNN. We tried FedMAML which is the MAML~\cite{maml} in the federated setting.
\end{itemize}

Due to the space limitation, we only list the aggregated metrics and the per-client improvement ratio in the following sections. The detailed per-client performances are in the Appendix.

\subsection{Results Analysis on Graph-DC}
\begin{wraptable}[17]{r}{0.55\textwidth}
    \centering
    \begin{tabular}{  c r r}
    \toprule
     &Mean  & \multirow{2}{*}{Overall ($\%$)}\\ & test accuracy ($\%$)&\\
    \midrule
Isolated &	 $71.16\pm0.99$ & $-$\\
 \midrule
FedAvg&	$61.67\pm0.98$ & $-11.64\%$\\
FedAvg+FT&	$61.71 \pm 0.44$ & $-9.19\%$\\
FedProx&   $61.70\pm0.41$ &  $-15.66\%$\\
 \midrule
FedBN&	$76.86\pm1.02$ & $0.42\%$\\
FedBN+FT&	$77.91\pm0.20$ & $3.80\%$\\
Ditto&	$61.99 \pm 0.90$ &  $-11.69\%$\\
\midrule
FedMAML&	$78.06 \pm 0.45$ &$5.82\%$\\
\bottomrule
    \end{tabular}
    \caption{Experiment results on Graph-DC dataset. ``FT" stands for fine-tuning.}
    \label{tab:exp_results}
\end{wraptable}
 We run each baseline three times and report the mean test accuracy. The experimental results are presented in Table~\ref{tab:exp_results}. We gain more insights into the proposed task from these preliminary experimental results. 
 As our experimental results indicate, gaining performance improvements from the collaboration is nontrivial. The majority of standard federated learning methods are significantly outperformed by the "isolated" baseline, illustrating the difficulties introduced by heterogeneity. 
As expected, the majority of personalized (such as FebBN and Ditto) or meta-learning-based methods (FedMAML) outperform standard federated learning methods, demonstrating their effectiveness in managing heterogeneity. Another observation is that the performance can be further improved when we fine-tune each client-specific model (learned via personalized or meta-learning) on the corresponding dataset.
\begin{figure}[ht]
\centering
\subfloat [FedAvg.]{
\includegraphics[width=0.225\textwidth]{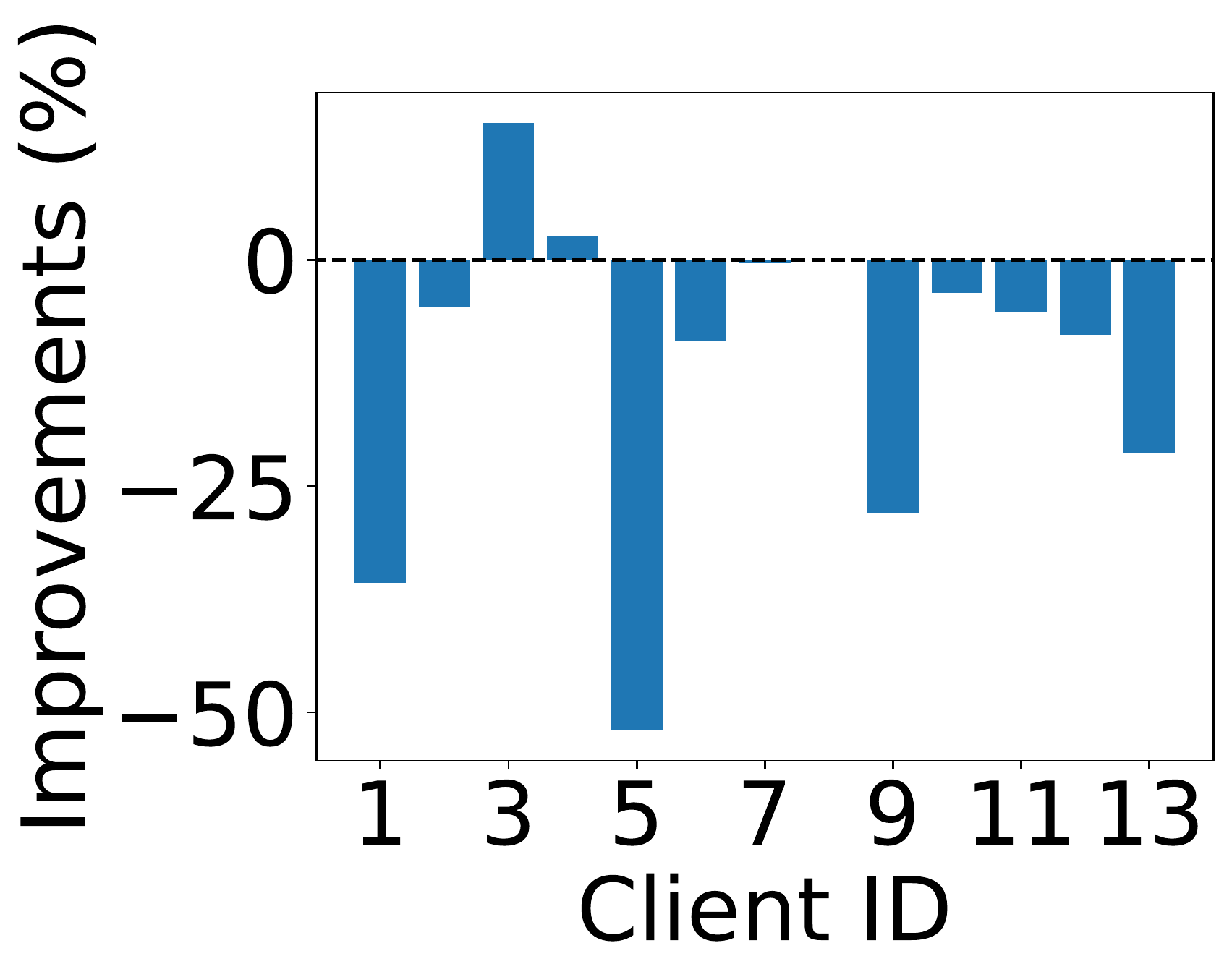}
\label{fig: acc_diff_fedavg}
}
\hfil
\vspace{-0.05in}
\subfloat [FedAvg+FT. ]{
\includegraphics[width=0.225\textwidth]{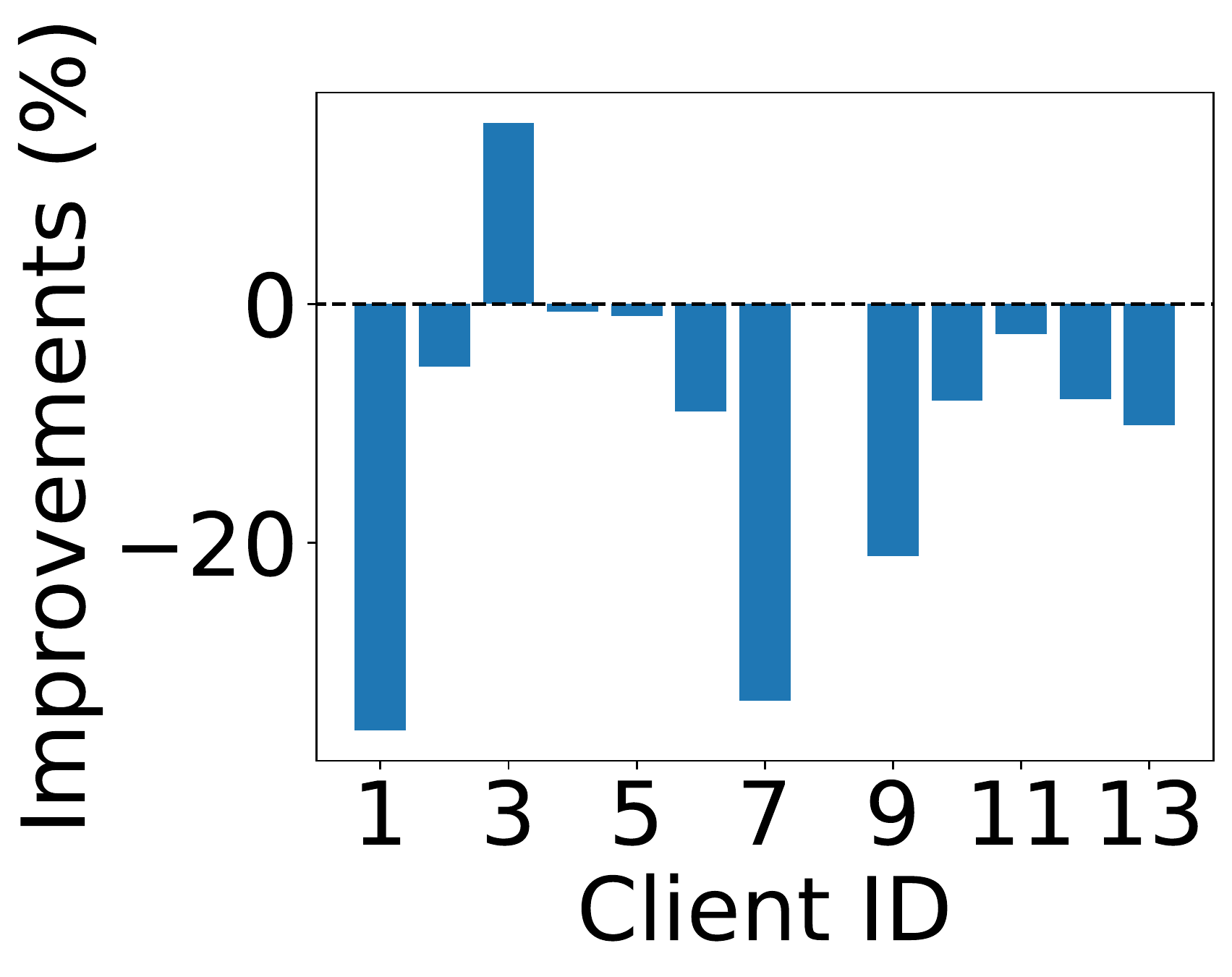}
\label{fig: acc_diff_fedavg_ft}
}
\hfil
\vspace{-0.05in}
\subfloat [FedProx.]{
\includegraphics[width=0.225\textwidth]{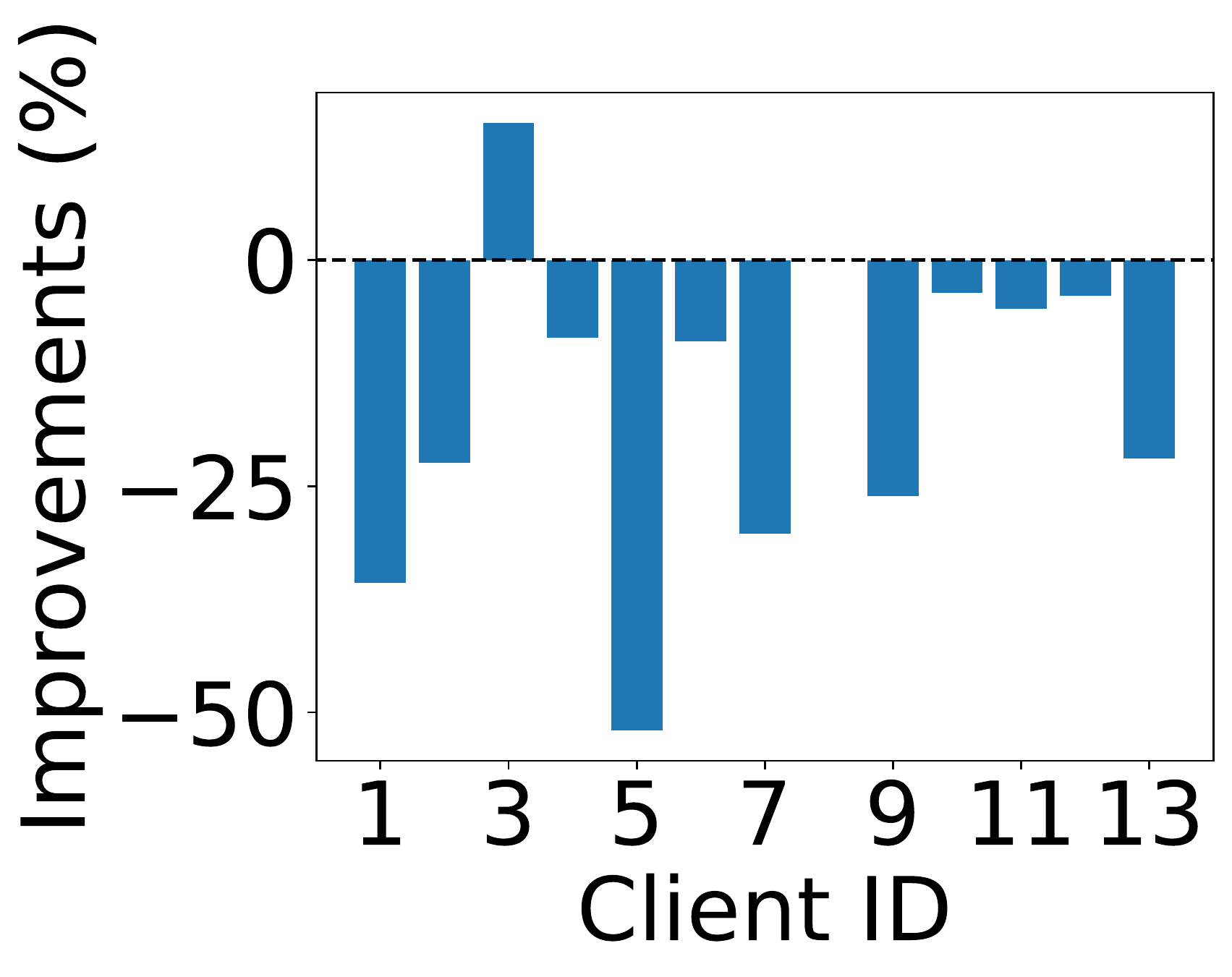}
\label{fig: acc_diff_fedprox}
}
\hfil
\subfloat [FedBN.]{
\includegraphics[width=0.225\textwidth]{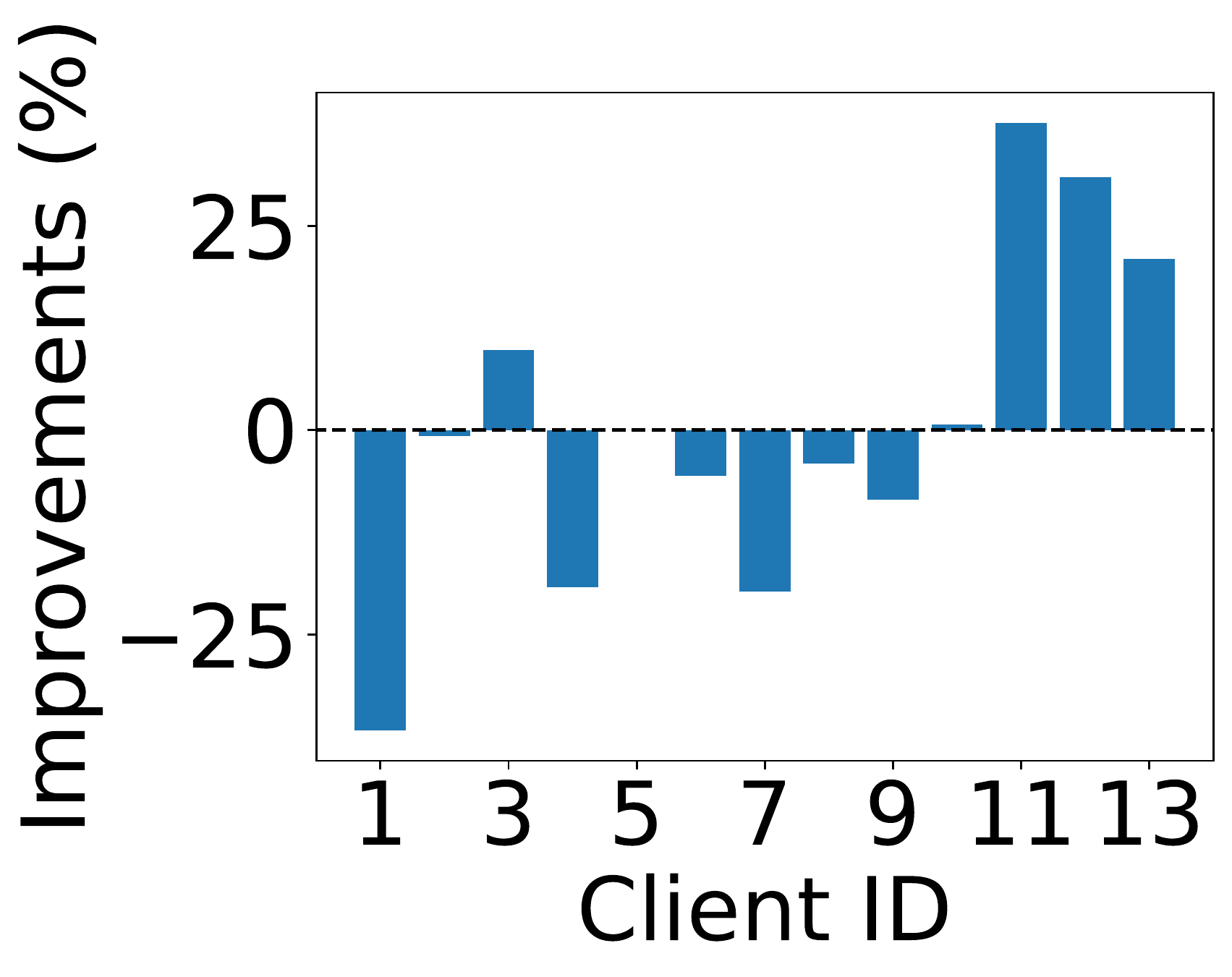}
\label{fig: acc_diff_fedbn}
}
\hfil
\subfloat [FedBN+FT.]{
\includegraphics[width=0.225\textwidth]{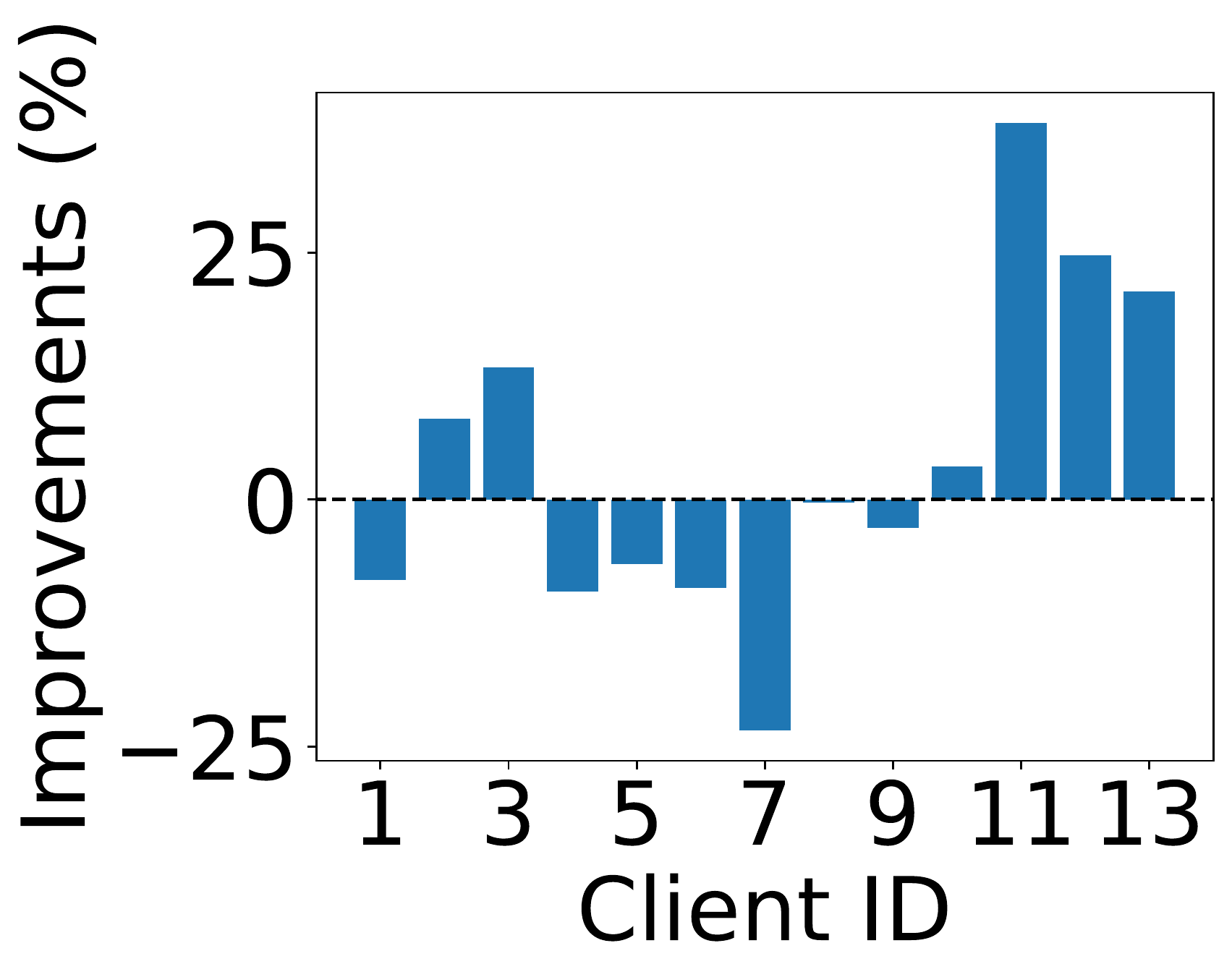}
\label{fig: acc_diff_fedbn_ft}
}
\hfil
\subfloat [Ditto.]{
\includegraphics[width=0.225\textwidth]{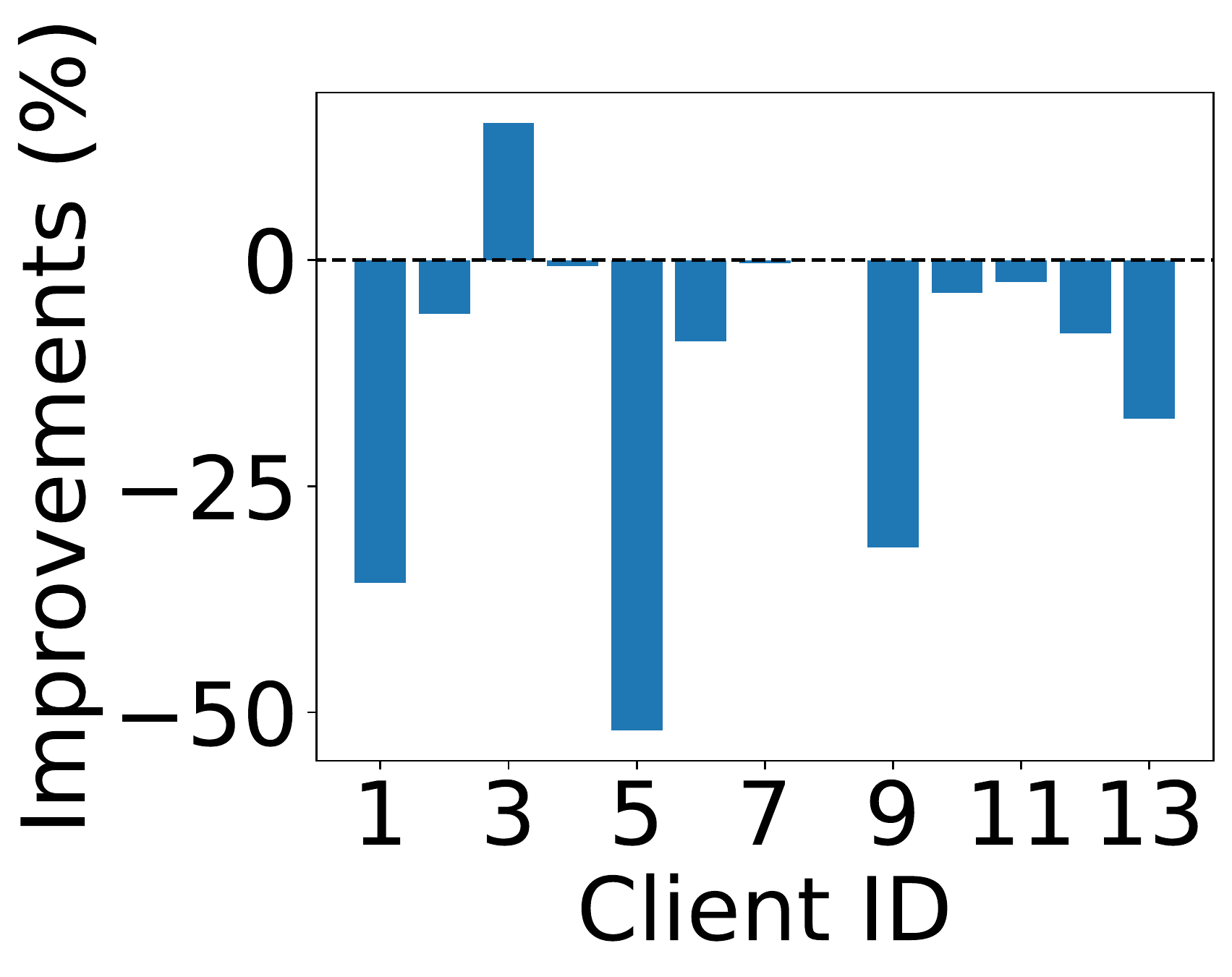}
\label{fig: acc_diff_ditto}
}
\hfil
\subfloat [FedMAML.]{
\includegraphics[width=0.225\textwidth]{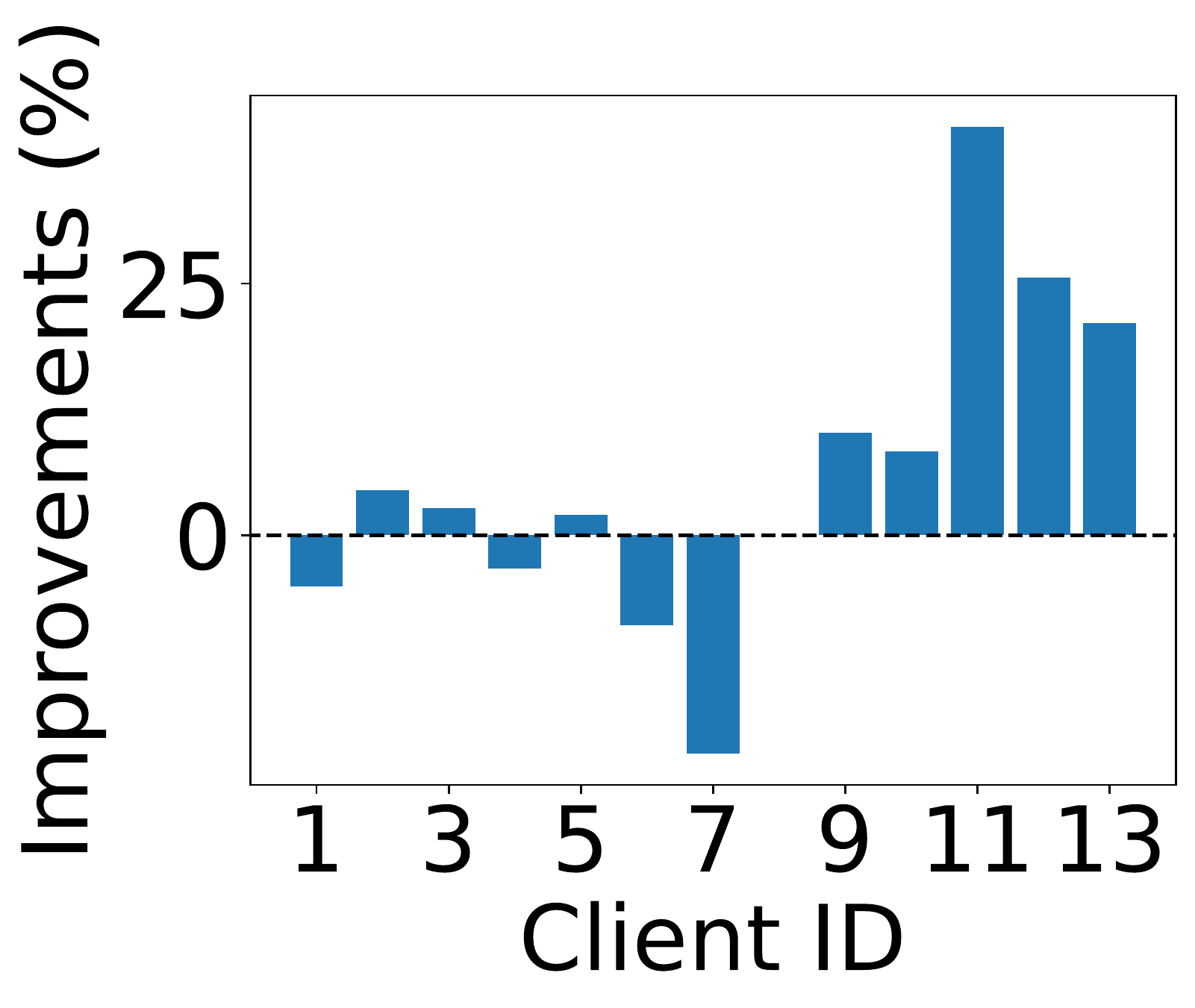}
\label{fig: acc_diff_maml}
}
\caption{Graph-DC: Per-client improvement ratio to Isolated.}
\label{fig: acc_diff}
\end{figure}

To further explore what will bring to each client by joining federated learning, we plot the per-client improvement ratio compared with the``isolated" method in Figure~\ref{fig: acc_diff}. From the figure, it is observed that federated learning method cannot consistently benefit all the clients, even in FedBN(figure~\ref{fig: acc_diff_fedbn}) and FedBN+FT(figure~\ref{fig: acc_diff_fedbn_ft}) and meta-learning based Fl method FedMAML (figure~\ref{fig: acc_diff_maml}) whose mean test accuracies are much better than ``isolated" method. As mentioned in section~\ref{sec:dataset}, the client IDs are numbered in ascending order of data sizes they own. Based on this numbering rule, we further observe the trend that clients with larger data sizes tend to gain more improvement from FL (shown in figure~\ref{fig: acc_diff_fedbn}, figure~\ref{fig: acc_diff_fedavg_ft} and figure~\ref{fig: acc_diff_maml}), and are less affected when joining FL hinders the overall performance. One possible reason is that methods such as FedAvg, MAML, and FedProx, the data size of each client are served as part of the weight when aggregating the model updates in the server, which results in that clients with more data may gain more from FL. This observation is also closely related to the fairness in federated learning~\cite{li2019fair}, which is left as our future work.

In a nutshell, on the one hand, our preliminary study above highlights the challenges inherent in our proposed heterogeneous federated learning. On the other hand, this demonstrates the enormous potential of contemporary federated learning methods.

\subsection{Results Analysis on Graph-DT}
\begin{wraptable}[15]{r}{0.4\textwidth}
    \centering
    \begin{tabular}{c r}
    \toprule
    & Overall ($\%$)\\
    \midrule
FedAvg & $1.69\%$ \\
FedAvg+FT & $2.13\%$ \\
FedProx & $-3.99\%$ \\
 \midrule
FedBN & $5.93\%$ \\
FedBN+FT & $8.48\%$ \\
Ditto & $-5.41\%$ \\
\midrule
FedMAML & $4.64\%$ \\
\bottomrule
    \end{tabular}
    \caption{Experiment results on Graph-DT dataset. ``FT" stands for fine-tuning.}
    \label{tab:exp_results_graph_dt}
\end{wraptable}
The experimental results of the overall improvement ratio over the ``isolated" method on Graph-DT dataset are shown in table~\ref{tab:exp_results_graph_dt}. In this dataset, the clients with classification tasks are evaluated with accuracy, and with regression tasks are evaluated with mean square error (MSE). It is observed from table that the standard FL method FedAvg is slightly better than the isolated method indicating the benefit of collaboratively training. We also observe that it is challenging to deal with heterogeneity, which requires careful design of federated learning method, as FedProx and Ditto have a negative improvement ratio. Similar to the performance on Graph-DC, meta-learning based method FedMAML and personalized FL based method FedBN continues to have a better performance indicating their potential to deal with heterogeneity. We also notice that after we fine-tune the FedBN, it eventually has the best performance, indicating fine-tuning may be a necessary step after the federated training.

\begin{figure}[t]
\centering
\subfloat [FedAvg.]{
\includegraphics[width=0.225\textwidth]{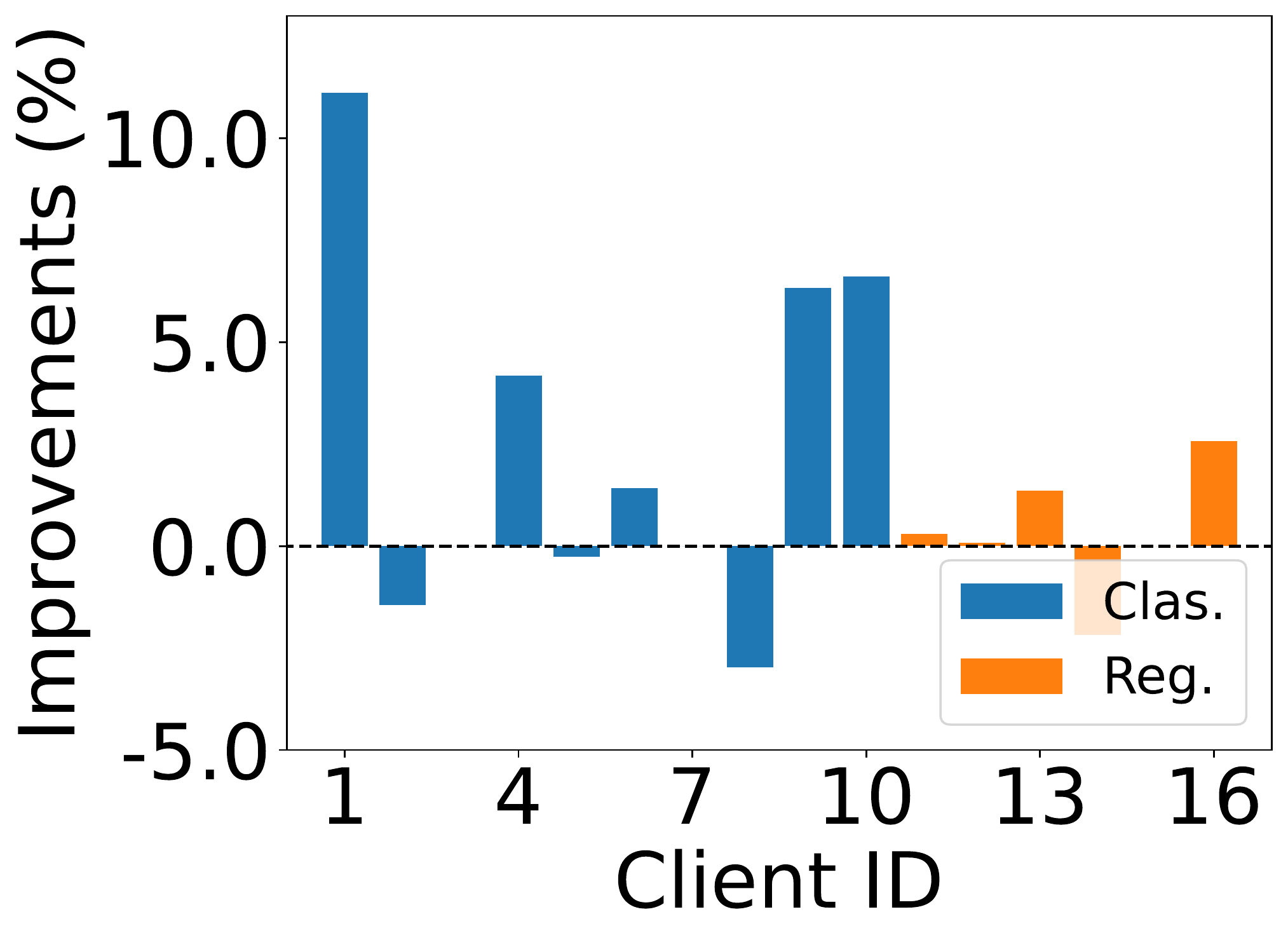}
\label{fig: imp_diff_fedavg}
}
\hfil
\subfloat [FedAvg+FT. ]{
\includegraphics[width=0.225\textwidth]{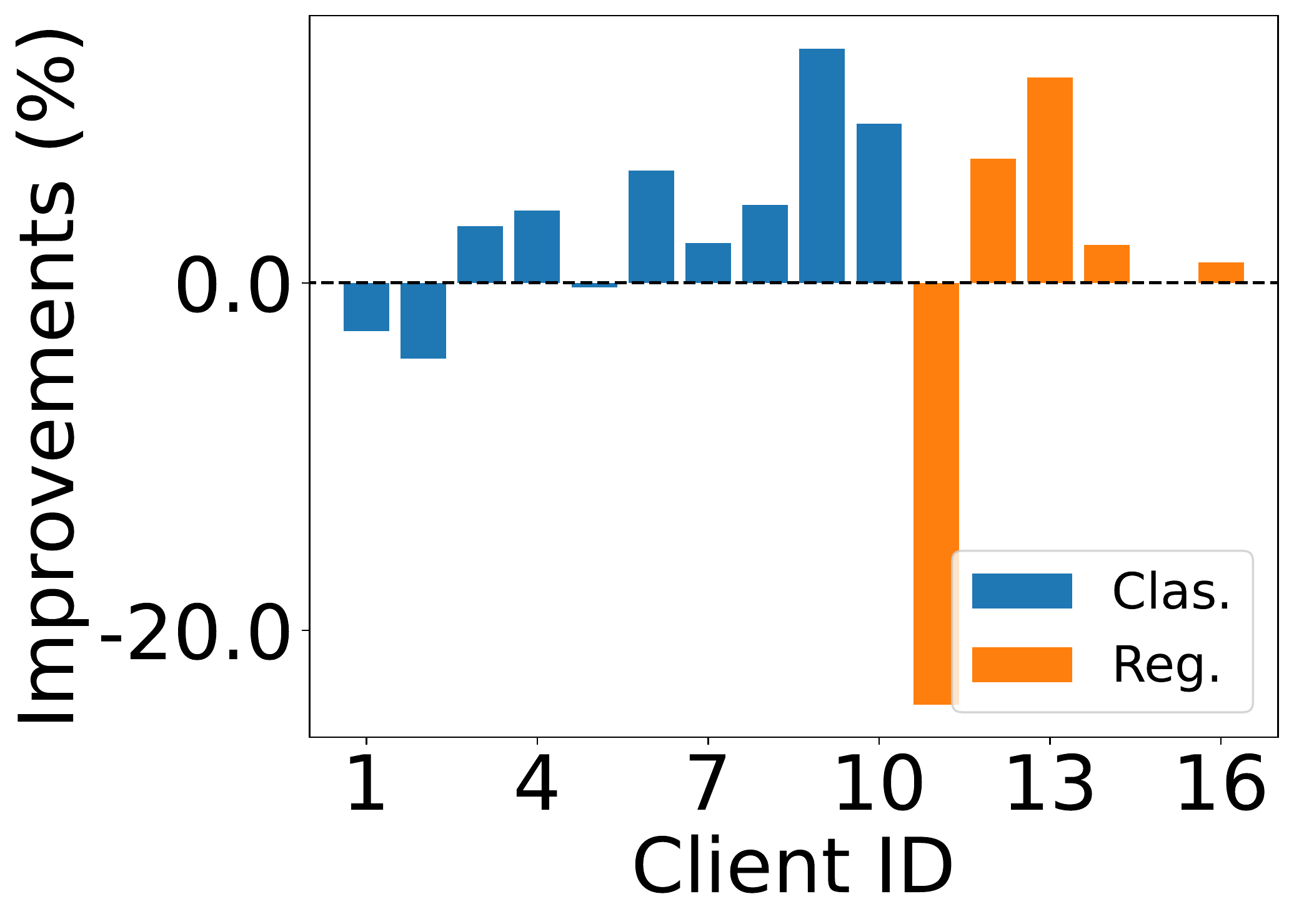}
\label{fig: imp_diff_fedavg_ft}
}
\hfil
\subfloat [FedProx.]{
\includegraphics[width=0.225\textwidth]{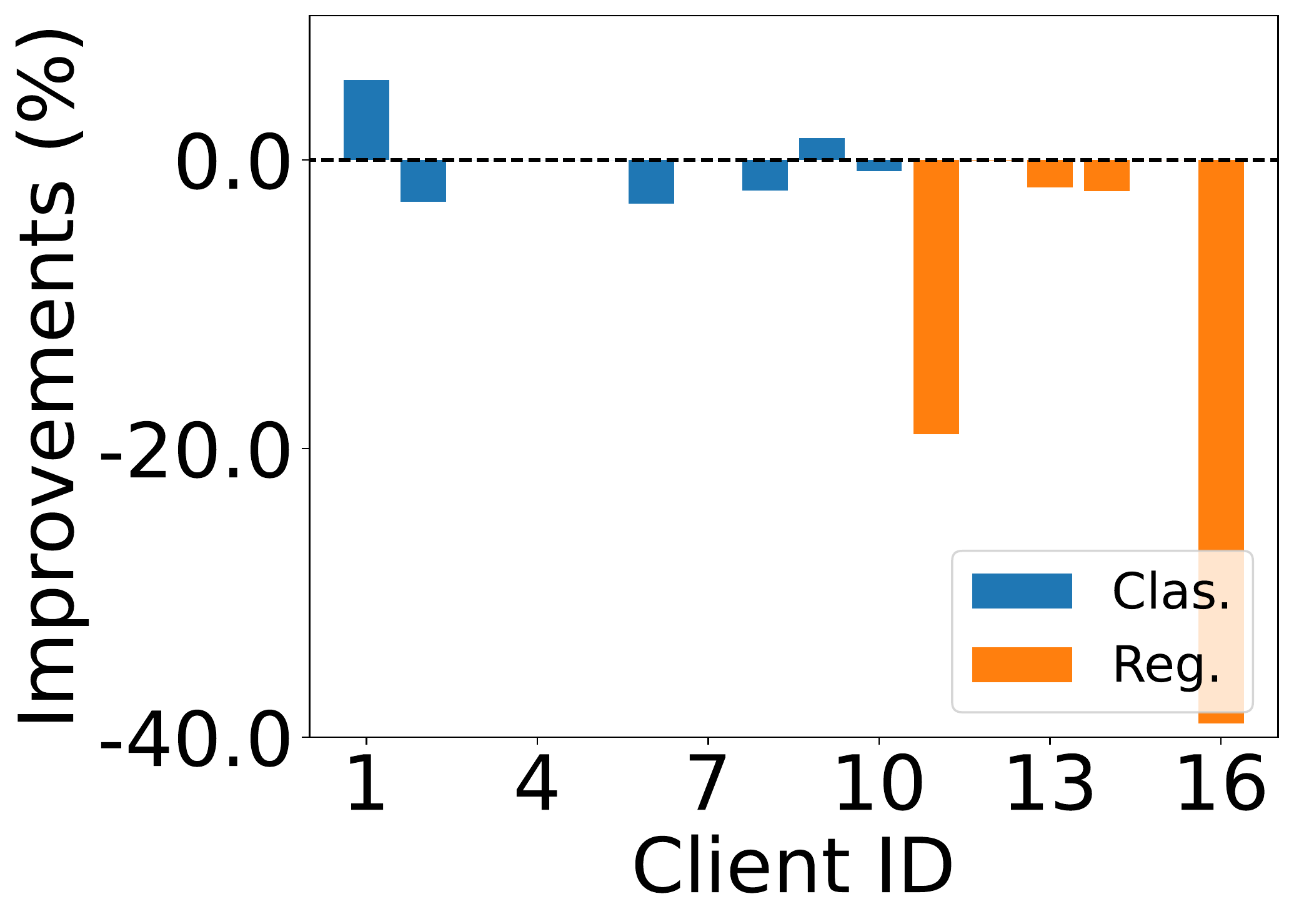}
\label{fig: imp_diff_fedprox}
}
\hfil
\subfloat [FedBN.]{
\includegraphics[width=0.225\textwidth]{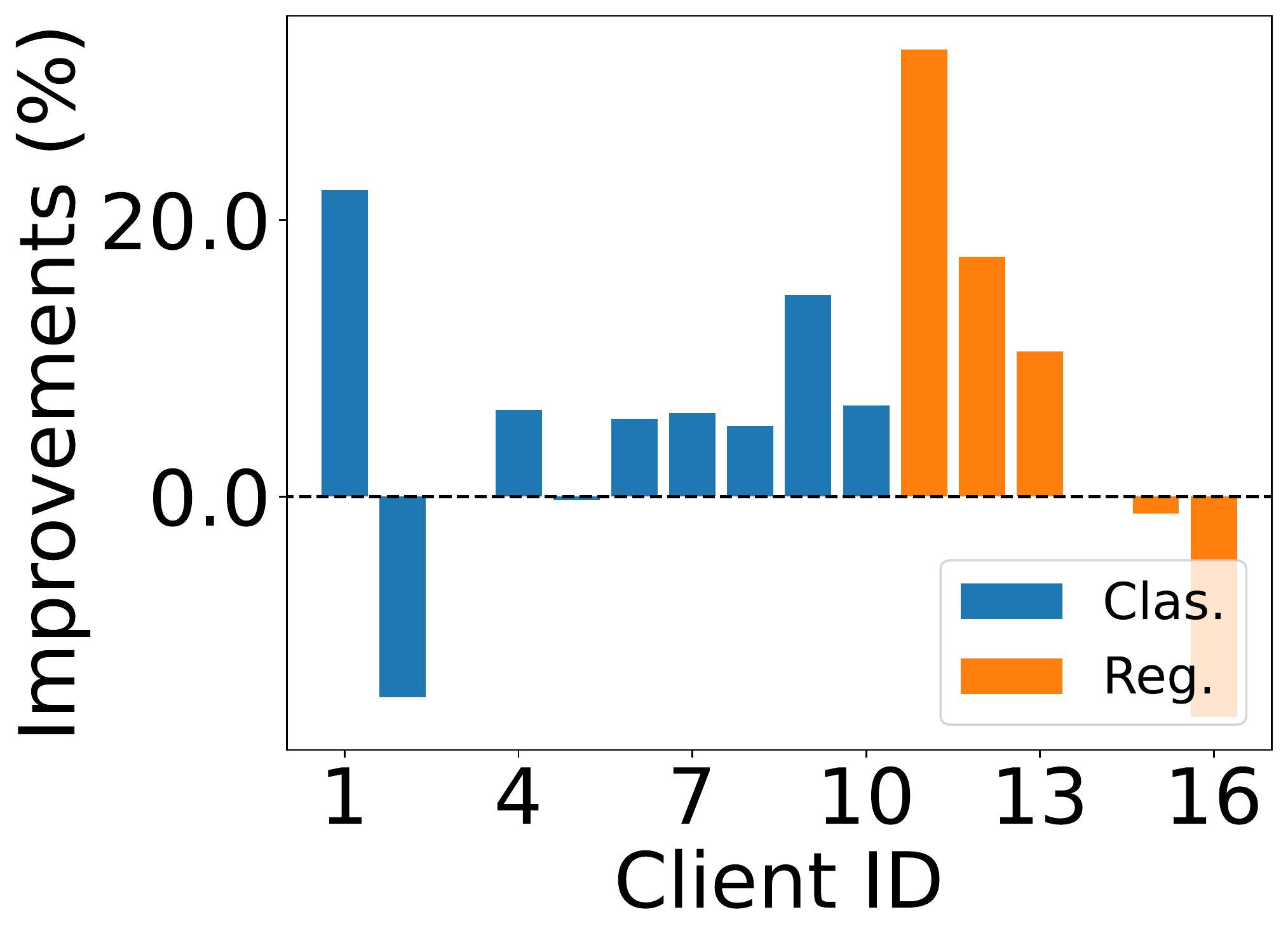}
\label{fig: imp_diff_fedbn}
}
\hfil
\subfloat [FedBN+FT.]{
\includegraphics[width=0.225\textwidth]{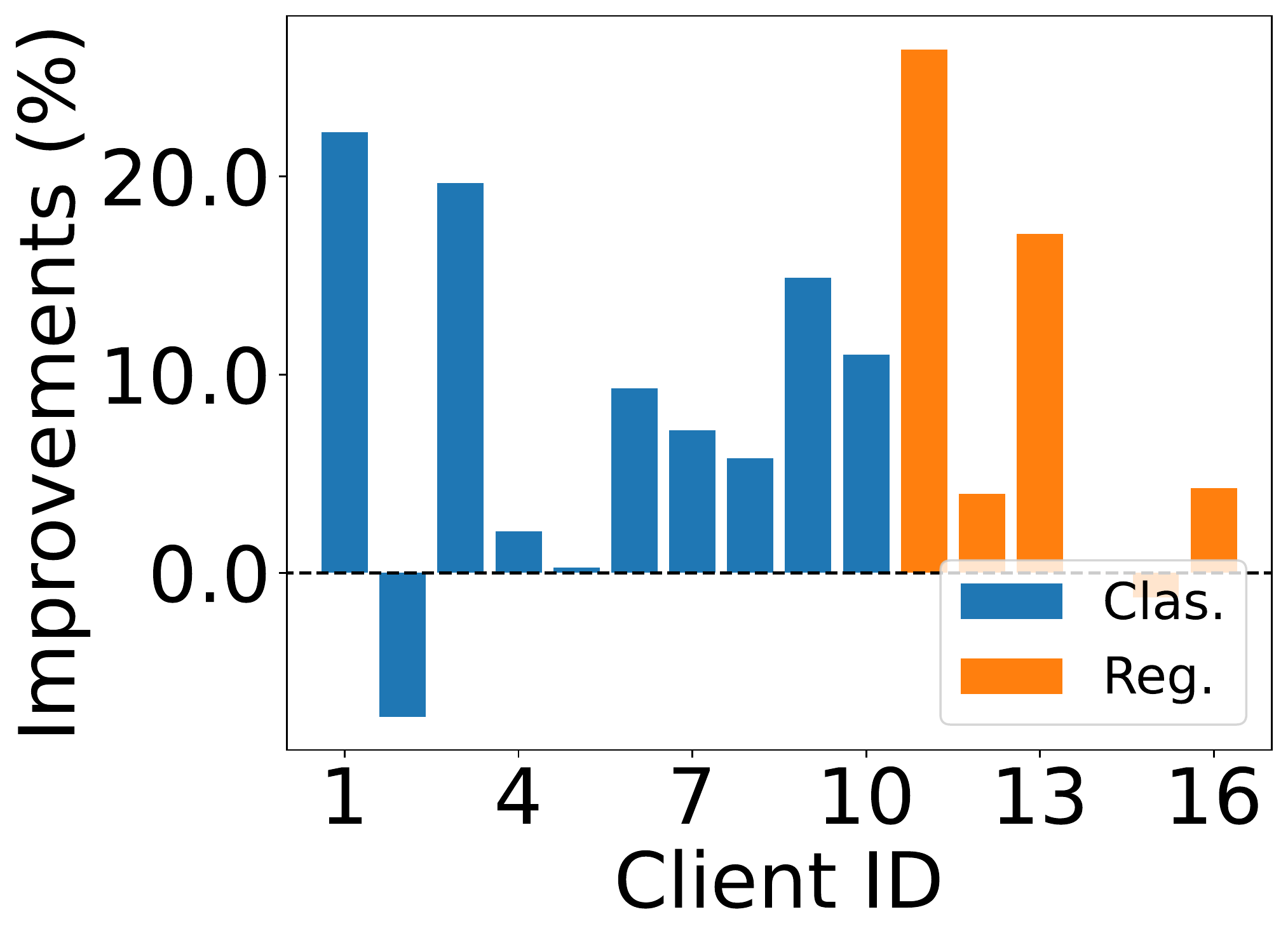}
\label{fig: imp_diff_fedbn_ft}
}
\hfil
\subfloat [Ditto.]{
\includegraphics[width=0.225\textwidth]{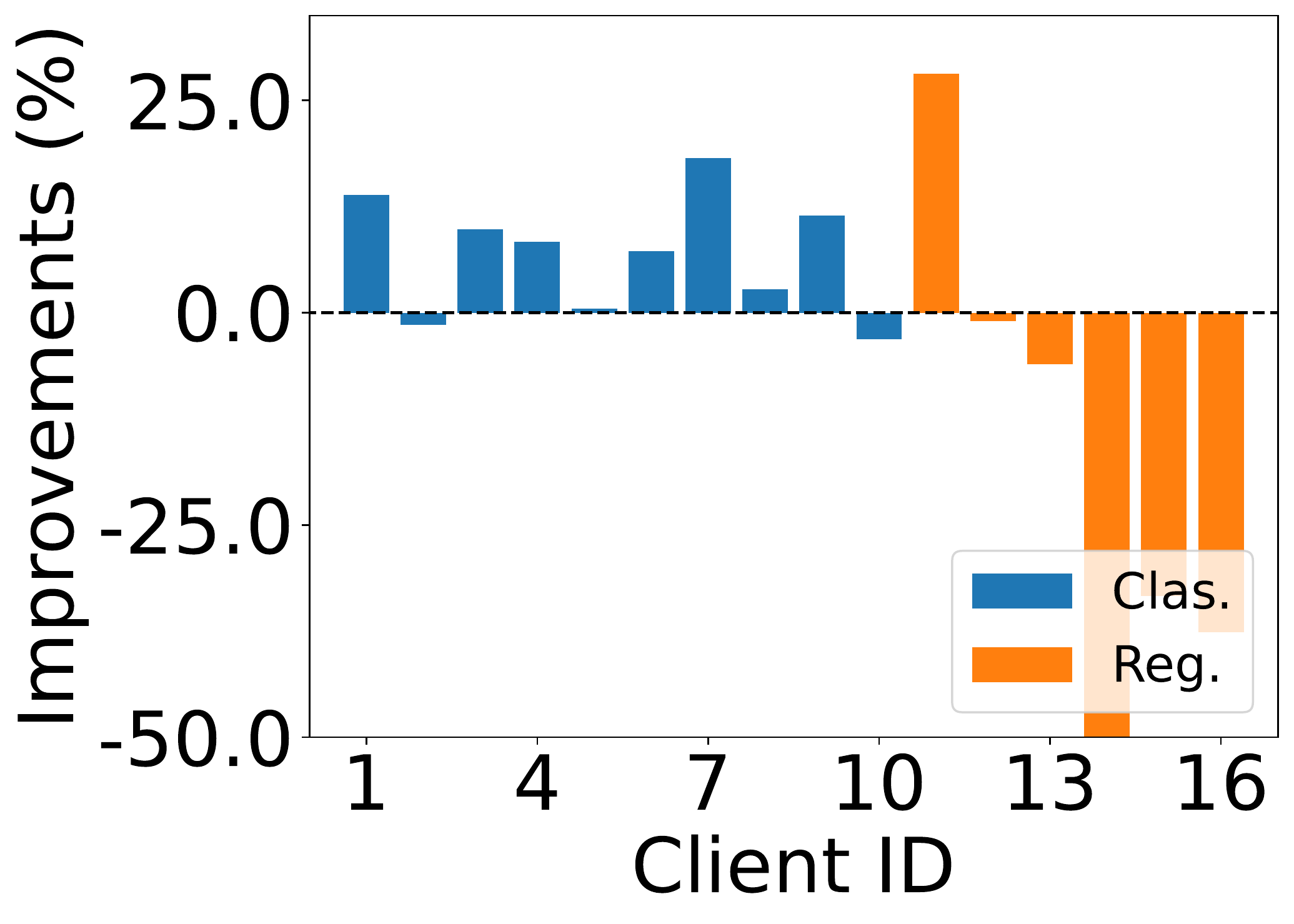}
\label{fig: imp_diff_ditto}
}
\hfil
\subfloat [FedMAML.]{
\includegraphics[width=0.225\textwidth]{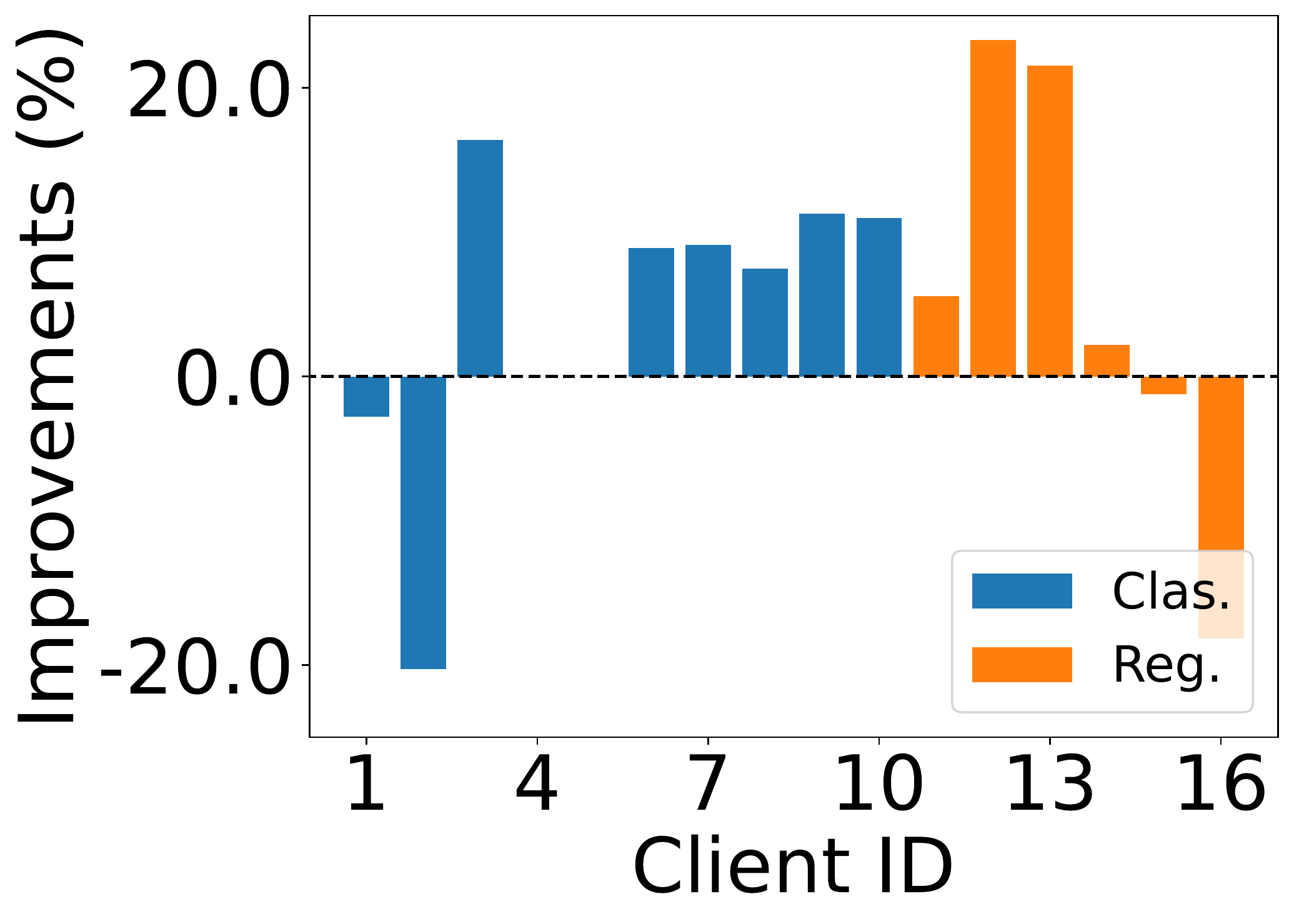}
\label{fig: imp_diff_maml}
}
\caption{Graph-DT: Per-client improvement ratio to Isolated.}
\label{fig: imp_diff}
\end{figure}

Next, we dive into the per-client performance to gain more insights into the included methods. Figure~\ref{fig: imp_diff} lists the per-client improvement ratio, and in each subgraph, the blue bar represents the clients with classification tasks and the orange bar represents clients with regression tasks. As the clients with the same task type are numbered in the ascending order of their dataset size, it is observed that the clients with small datasets are able to borrow strength from joining the federated learning in the personalized federated learning based (figure~\ref{fig: acc_diff_fedbn}) and meta-learning based method (figure~\ref{fig: acc_diff_maml}). This promising trend indicates the great potential of personalized FL and metal-learning based methods in federated hetero-task learning.

\subsection{Results Analysis on Text-DT}
\begin{wraptable}[14]{r}{0.6\textwidth}
    \centering
    \begin{tabular}{ c r r r r}
    \toprule
    & Overall ($\%$)&Client 1 & Client 2& Client 3\\
 \midrule
FedAvg & $2.22\%$ & $1.03\%$ & $3.31\%$ & $2.32\%$ \\
FedAvg+FT & $2.47\%$& $1.05\%$ & $3.29\%$ & $3.08\%$ \\
FedProx & $2.32\%$ & $0.95\%$ & $3.38\%$ & $2.62\%$ \\
 \midrule
FedBN & $2.41\%$ & $0.99\%$ & $3.32\%$ & $2.92\%$ \\
FedBN+FT & $2.51\%$& $1.00\%$ & $3.31\%$ & $3.21\%$ \\
Ditto & $3.53\%$ & $0.73\%$ & $3.21\%$ & $6.66\%$ \\
\midrule
FedMAML & $3.57\%$& $1.07\%$ & $2.40\%$ & $7.24\%$ \\
\bottomrule
    \end{tabular}
    \caption{Experiment results on Text-DT dataset. ``FT" stands for fine-tuning.}
    \label{tab:exp_results_text_dt}
\end{wraptable}
On the Text-DT dataset, the evaluation metric on client 1 is Pearson correlation, on client 2 is accuracy and on client 3 is exact match.
The overall improvement ratio and the per-client improvement ratio compared to the ``isolated" are shown in table~\ref{tab:exp_results_text_dt}. From the table, it is observed that all federated learning methods bring improvement to each client's task. Especially, the personalization based (FedBN and Ditto) and the meta learning based (FedMAML) federated learning methods are slightly better than the standard federated learning methods (FedAvg and FedProx), which reflects the potential of managing heterogeneity in performance improvement.

%% file: subfile/5_Conclusion.tex
\section{Conclusion}
\label{sec:conclusion}
With growing concerns about privacy, federated learning has become a popular solution for privacy protection in machine learning applications. In the real-world scenario, it is often the case that there exists both the data heterogeneity and the learning goal heterogeneity among the different clients. To enlarge the application scope of federated learning, in this paper, we generalize the classic federated learning to federated hetero-task learning, which considers the two kinds of heterogeneity. Federated hetero-task learning is closely related to several important domains such as personalized federated learning, multi-task learning, meta-learning, and so on. To promote the development of the federated hetero-task learning and facilitate the interdisciplinary research, we propose B-FHTL, an easy-to-use federated hetero-task learning benchmark, including the federated dataset, the FL protocols, and the evaluation mechanism. The federated dataset reflects heterogeneous data and learning goals across different clients; the built-in necessary federated learning protocols ensure the convenient method implementation and fair comparisons. The preliminary experiment analysis is given to gain insights into the federated hetero-task learning by conducting the experiments on the provided federated datasets. 
Overall, our benchmark gives a comprehensive simulation of the federated hetero-task learning setting, which facilitates the development of innovative methods and encourages interdisciplinary research.

\textbf{Discussion.} 
With the development of federated hetero-task learning, more innovative methods solving the heterogeneity challenge 
will be proposed in the literature. We will keep on including the state-of-the-art federated hetero-task learning methods and meanwhile welcome the contributions to our benchmark. Furthermore,  the current three datasets are from graph and text domains, and we will include more federated datasets across different domains, including CV, speech, healthcare, and so on.

%% file: subfile/appendix.tex
\section*{Appendix}
In appendix, we provide more details about our benchmark and the experimental results.
\section{Implementation}

\subsection{Graph-DC} 
\textbf{Model}. 
To make a fair comparison, we adopt Graph Isomorphism Network (GIN) as the basic neural network for all baselines, which is a two-layer message-passing GNN model with an MLP layer.

\textbf{Hyper-parameters.} 
The hyper-parameters of the chosen baselines are shown in the Table~\ref{tab:hyper1}. 

\begin{table}[ht]
    \centering
    \begin{tabular}{ccc}
    \toprule
    Baseline & Hyper-parameter &  Value\\
    \midrule
    \multirow{3}{*}{FedBN}& learning rate     &  $0.05$\\
    & batch size & $64$ \\
    &local training steps & $4$ \\
    \midrule
    \multirow{3}{*}{FedBN+FT}& learning rate     &  $0.5$\\
    & batch size & $64$ \\
    &local training steps & $10$ \\
    \midrule
    \multirow{4}{*}{Ditto}& learning rate     &  $0.005$\\
    & batch size & $64$ \\
    &local training steps & $4$ \\
    & personalization regularization weight & $0.1$\\ 
    \midrule
    \multirow{3}{*}{FedAvg} & learning rate & 0.5\\
    & batch size & $64$ \\
    & local training steps & $2$ \\
    \midrule
    \multirow{4}{*}{FedAvg+FT} & learning rate & 0.1\\
    & batch size & $64$ \\
    & local training steps & $4$ \\
    & fine tuning steps & $5$ \\
    \midrule
    \multirow{3}{*}{FedProx} & learning rate & 0.5\\
    & batch size & $64$ \\
    & local training steps & $4$ \\
    \midrule
    \multirow{4}{*}{FedMAML} & outer learning rate & 0.01\\
    & inner learning rate & 0.01\\
    & batch size & $64$ \\
    & local training steps & 1 \\
    \bottomrule
    \end{tabular}
    \caption{Hyper-parameters of the baselines on Graph-DC }
    \label{tab:hyper1}
\end{table}

\subsection{Graph-DT}

\begin{table}[th]
    \centering
    \begin{tabular}{ccc}
    \toprule
         Baseline & Hyper-parameter & Value  \\ 
         \hline
         \multirow{4}{*}{FedBN} & \multirow{2}{*}{learning rate} & [0.1, 0.05, 0.0001, 0.05, 0.1, 0.025, 0.01, 0.025,  \\
         & & 0.05, 0.05, 0.1, 0.01, 0.05, 0.05, 0.05, 0.1] \\
         & batch size & 64 \\ 
         & local training steps & 21 \\
         \hline
         \multirow{4}{*}{FedBN+FT} & \multirow{2}{*}{learning rate} & [0.1, 0.05, 0.0001, 0.05, 0.1, 0.025, 0.01, 0.025,  \\
         & & 0.05, 0.05, 0.1, 0.01, 0.05, 0.05, 0.05, 0.1] \\
         & batch size & 64 \\ 
          & local training steps & 21 \\
         \hline
         \multirow{5}{*}{Ditto} & \multirow{2}{*}{learning rate} & [0.1, 0.05, 0.0001, 0.05, 0.1, 0.05, 0.01, 0.05,   \\
         & & 0.05, 0.05, 0.1, 0.01, 0.1, 0.05, 0.05, 0.1] \\
         & batch size & 64 \\ 
         & local training steps & 21 \\
         & personalization regularization weight & 0.01 \\ 
         \hline
         \multirow{4}{*}{FedAvg} & \multirow{2}{*}{learning rate} & [0.1, 0.05, 0.0001, 0.05, 0.1, 0.05, 0.01, 0.05,   \\
         & & 0.05, 0.05, 0.1, 0.01, 0.1, 0.05, 0.05, 0.1] \\
         & batch size & 64 \\ 
         & local training steps & 21 \\
         \hline
         \multirow{4}{*}{FedAvg+FT} & \multirow{2}{*}{learning rate} & [0.1, 0.05, 0.0001, 0.05, 0.1, 0.05, 0.01, 0.05,   \\
         & & 0.05, 0.05, 0.1, 0.01, 0.1, 0.05, 0.05, 0.1] \\
         & batch size & 64 \\ 
         & local training steps & 21 \\
         \hline
         \multirow{4}{*}{FedProx} & \multirow{2}{*}{learning rate} & [0.1, 0.05, 0.0001, 0.05, 0.1, 0.05, 0.01, 0.05,   \\
         & & 0.05, 0.05, 0.1, 0.01, 0.1, 0.05, 0.05, 0.1] \\
         & batch size & 64 \\ 
         & local training steps & 21 \\
         \hline
         \multirow{6}{*}{FedMAML} & \multirow{2}{*}{outer learning rate} & [0.1, 0.05, 0.0001, 0.05, 0.1, 0.05, 0.01, 0.05,   \\
         & & 0.05, 0.05, 0.1, 0.01, 0.1, 0.05, 0.05, 0.1] \\
         & \multirow{2}{*}{inner learning rate} & [0.005, 0.1, 0.01, 0.1, 0.05, 0.1, 0.0001, 0.1,   \\
         & & 0.001, 0.005, 0.1, 0.1, 0.1, 0.005, 0.05, 0.001] \\
         & batch size & 64 \\ 
         & local training steps & 21 \\
    \bottomrule
    \end{tabular}
    \caption{Hyper-parameters of the baselines on Graph-DT}
    \label{tab:hyper2}
\end{table}

\textbf{Model}.
We also choose the same GIN model for the dataset Graph-DT. 
Considering the dataset contains both regression and classification tasks, we modify the dimension of the output layer to fit the given task. 

\textbf{Pre-processing for features}.
As stated before, Graph-DT is consisted of the datasets chosen from TUDataset and Moluculenet. 
For the dataset from Moluculenet, we encode the features as follows
\begin{itemize}
\item Node Featurizer
    \begin{itemize}
    \item One hot encoding of the atom type. The supported atom types include 'B', 'C', 'N', 'O', 'S', 'F', 'Si', 'P', 'Cl', 'As', 'Se', 'Br', 'Te', 'I', 'At', 'other'.
    \item One hot encoding of the atom hybridization. The supported possibilities include 'SP', 'SP2', 'SP3', 'SP3D', 'SP3D2', 'other'.
    \end{itemize}
\item Edge Featurizer
    \begin{itemize}
    \item One hot encoding of the bond type. The supported bond types include 'SINGLE', 'DOUBLE', 'TRIPLE', 'AROMATIC'.
    \item One hot encoding of the stereo configuration of a bond. The supported bond stereo configurations include 'STEREONONE', 'STEREOANY', 'STEREOZ', 'STEREOE'.
    \end{itemize}
\end{itemize}

\textbf{Pre-processing for Labels}. 
To obtain a more appropriate target distribution for the regression tasks, we transform the target values with logarithmic function. 
For the dataset FreeSolv, we use $y = \log(-y + 5)$.
For the dataset ZINC\_full, $y = \log(-y + 5)$.
When we consider the multi regression dataset alchemy\_full, for label 3, 5, 7, 8, we use $y = \log(-y)$, while for label 10, 11, we finally transform the target by $y = \log(y)$. 
For multi regression dataset QM9, $y = \log(y)$ for label 0, 16, 17 and 18.

\textbf{Hyper-parameters}. 
With highly heterogeneous tasks in Graph-DT, we set client-wise hyper-parameters with the support of FederatedScope. 
The detailed hyper-parameters are shown in Table\ref{tab:hyper2}.

\subsection{Text-DT}
\textbf{Model}.
For Text-DT, we use BERT base as our training model for all clients, and adjust the output layer to fit different tasks. 
Considering the training of FedMAML involves the computation of hessian matrix, we only share the last linear layer within the encoder and the output layer during federated training. 
While for other baselines, we share the whole model except the output layer. 

\textbf{Hyper-parameters}.
We present the hyper-parameters for the baselines with Text-DT in Table~\ref{tab:hyper3}

\begin{table}[t]
    \centering
    \begin{tabular}{ccc}
    \toprule
    Baseline & Hyper-parameter &  Value\\
    \midrule
    \multirow{3}{*}{FedBN}& learning rate     &  3e-5\\
    & batch size & $32$ \\
    &local training steps & $500$ \\
    \midrule
    \multirow{3}{*}{FedBN+FT}& learning rate     &  3e-6\\
    & batch size & $32$ \\
    & local training steps & $500$ \\
    \midrule
    \multirow{4}{*}{Ditto}& learning rate     &  3e-5\\
    & batch size & $32$ \\
    &local training steps & $250$ \\
    & personalization regularization weight & 0.001\\ 
    \midrule
    \multirow{3}{*}{FedAvg} & learning rate & 3e-5\\
    & batch size & $32$ \\
    & local training steps & $500$ \\
    \midrule
    \multirow{3}{*}{FedAvg+FT} & learning rate & 3e-6\\
    & batch size & $32$ \\
    & local training steps & $500$ \\
    \midrule
    \multirow{3}{*}{FedProx} & learning rate & 3e-5\\
    & batch size & $32$ \\
    & local training steps & $500$ \\
    \midrule
    \multirow{4}{*}{FedMAML} & outer learning rate & 1e-3\\
    & inner learning rate & 1e-3\\
    & batch size & $32$ \\
    & local training steps & 50 \\
    \bottomrule
    \end{tabular}
    \caption{Hyper-parameters of the baselines on Text-DT }
    \label{tab:hyper3}
\end{table}

\section{Per-client Results}
To compare the baselines with different clients, we also present the per-client performance for Graph-DC, Graph-DT and Text-DT in Table~\ref{tab:per_client1}, Table~\ref{tab:per_client2} and Table~\ref{tab:per_client3} respectively. 

\begin{table}[th]
    \centering
    \resizebox{0.95\textwidth}{!}{
    \begin{tabular}{c c c c c c c c c}
    \toprule
Client ID & Isolated&FedAvg&FedAvg+FT&FedProx&FedBN&FedBN+FT&Ditto&FedMAML\\ 
\midrule
 1 & $0.860 \pm 0.033$&$0.553 \pm 0.015$&$0.553 \pm 0.065$&$0.544 \pm 0.118$&$0.789 \pm 0.021$&$0.553 \pm 0.030$&$0.553 \pm 0.030$&$0.816 \pm 0.037$\\ 
 2 & $0.657 \pm 0.123$&$0.623 \pm 0.007$&$0.623 \pm 0.007$&$0.652 \pm 0.073$&$0.711 \pm 0.007$&$0.618 \pm 0.057$&$0.510 \pm 0.076$&$0.686 \pm 0.018$\\ 
 3 & $0.541 \pm 0.018$&$0.623 \pm 0.063$&$0.623 \pm 0.034$&$0.594 \pm 0.043$&$0.614 \pm 0.056$&$0.623 \pm 0.113$&$0.623 \pm 0.080$&$0.556 \pm 0.007$\\ 
 4 & $0.709 \pm 0.029$&$0.728 \pm 0.018$&$0.704 \pm 0.065$&$0.573 \pm 0.178$&$0.643 \pm 0.087$&$0.704 \pm 0.065$&$0.648 \pm 0.100$&$0.685 \pm 0.018$\\ 
 5 & $0.823 \pm 0.006$&$0.395 \pm 0.104$&$0.815 \pm 0.192$&$0.823 \pm 0.006$&$0.770 \pm 0.051$&$0.395 \pm 0.031$&$0.395 \pm 0.098$&$0.840 \pm 0.010$\\ 
 6 & $0.830 \pm 0.023$&$0.755 \pm 0.163$&$0.755 \pm 0.145$&$0.784 \pm 0.018$&$0.755 \pm 0.034$&$0.755 \pm 0.090$&$0.755 \pm 0.034$&$0.755 \pm 0.045$\\ 
 7 & $0.667 \pm 0.035$&$0.664 \pm 0.075$&$0.445 \pm 0.155$&$0.535 \pm 0.127$&$0.511 \pm 0.075$&$0.664 \pm 0.091$&$0.465 \pm 0.143$&$0.522 \pm 0.121$\\ 
 8 & $0.985 \pm 0.071$&$0.985 \pm 0.034$&$0.985 \pm 0.103$&$0.945 \pm 0.032$&$0.982 \pm 0.005$&$0.985 \pm 0.123$&$0.985 \pm 0.033$&$0.985 \pm 0.055$\\ 
 9 & $0.703 \pm 0.037$&$0.507 \pm 0.015$&$0.555 \pm 0.029$&$0.643 \pm 0.049$&$0.683 \pm 0.025$&$0.480 \pm 0.007$&$0.520 \pm 0.011$&$0.775 \pm 0.018$\\ 
 10 & $0.853 \pm 0.019$&$0.823 \pm 0.035$&$0.784 \pm 0.054$&$0.859 \pm 0.027$&$0.882 \pm 0.031$&$0.823 \pm 0.054$&$0.823 \pm 0.064$&$0.924 \pm 0.015$\\ 
 11 & $0.531 \pm 0.026$&$0.500 \pm 0.002$&$0.517 \pm 0.015$&$0.730 \pm 0.017$&$0.733 \pm 0.009$&$0.518 \pm 0.012$&$0.502 \pm 0.003$&$0.746 \pm 0.014$\\ 
 12 & $0.567 \pm 0.043$&$0.520 \pm 0.001$&$0.522 \pm 0.002$&$0.743 \pm 0.012$&$0.707 \pm 0.042$&$0.521 \pm 0.001$&$0.545 \pm 0.037$&$0.712 \pm 0.011$\\ 
 13 & $0.663 \pm 0.091$&$0.522 \pm 0.102$&$0.596 \pm 0.102$&$0.802 \pm 0.011$&$0.803 \pm 0.006$&$0.546 \pm 0.039$&$0.518 \pm 0.096$&$0.802 \pm 0.007$\\
 \bottomrule
    \end{tabular}}
    \caption{Per-client results on Graph-DC.}
    \label{tab:per_client1}
\end{table}

\begin{table}[th]
    \centering
    \resizebox{0.95\textwidth}{!}{
    \begin{tabular}{c c c c c c c c c}
    \toprule
Client ID & Isolated&FedAvg&FedAvg+FT&FedProx&FedBN&FedBN+FT&Ditto&FedMAML\\ 
\midrule
 1 & $0.632 \pm 0.091$&$0.702 \pm 0.121$&$0.614 \pm 0.133$&$0.667 \pm 0.110$&$0.772 \pm 0.121$&$0.772 \pm 0.030$&$0.719 \pm 0.080$&$0.614 \pm 0.030$\\ 
 2 & $0.676 \pm 0.029$&$0.667 \pm 0.045$&$0.647 \pm 0.029$&$0.657 \pm 0.034$&$0.578 \pm 0.045$&$0.627 \pm 0.061$&$0.667 \pm 0.017$&$0.539 \pm 0.034$\\ 
 3 & $0.581 \pm 0.280$&$0.581 \pm 0.280$&$0.600 \pm 0.247$&$0.581 \pm 0.280$&$0.581 \pm 0.280$&$0.695 \pm 0.083$&$0.638 \pm 0.044$&$0.676 \pm 0.017$\\ 
 4 & $0.457 \pm 0.000$&$0.476 \pm 0.033$&$0.476 \pm 0.033$&$0.457 \pm 0.000$&$0.486 \pm 0.029$&$0.467 \pm 0.044$&$0.495 \pm 0.059$&$0.457 \pm 0.000$\\ 
 5 & $0.889 \pm 0.000$&$0.887 \pm 0.004$&$0.887 \pm 0.004$&$0.889 \pm 0.000$&$0.887 \pm 0.011$&$0.891 \pm 0.004$&$0.893 \pm 0.008$&$0.889 \pm 0.000$\\ 
 6 & $0.711 \pm 0.019$&$0.721 \pm 0.029$&$0.757 \pm 0.026$&$0.690 \pm 0.061$&$0.751 \pm 0.005$&$0.777 \pm 0.025$&$0.763 \pm 0.028$&$0.774 \pm 0.016$\\ 
 7 & $0.579 \pm 0.011$&$0.579 \pm 0.011$&$0.592 \pm 0.020$&$0.579 \pm 0.011$&$0.614 \pm 0.023$&$0.621 \pm 0.015$&$0.684 \pm 0.043$&$0.632 \pm 0.148$\\ 
 8 & $0.766 \pm 0.023$&$0.744 \pm 0.045$&$0.801 \pm 0.019$&$0.750 \pm 0.058$&$0.806 \pm 0.010$&$0.810 \pm 0.012$&$0.788 \pm 0.029$&$0.824 \pm 0.009$\\ 
 9 & $0.589 \pm 0.071$&$0.626 \pm 0.003$&$0.668 \pm 0.011$&$0.598 \pm 0.066$&$0.675 \pm 0.022$&$0.676 \pm 0.018$&$0.656 \pm 0.013$&$0.655 \pm 0.087$\\ 
 10 & $0.624 \pm 0.071$&$0.665 \pm 0.044$&$0.681 \pm 0.005$&$0.619 \pm 0.087$&$0.665 \pm 0.012$&$0.693 \pm 0.006$&$0.605 \pm 0.029$&$0.693 \pm 0.016$\\ 
 11 & $0.099 \pm 0.031$&$0.098 \pm 0.020$&$0.122 \pm 0.033$&$0.117 \pm 0.014$&$0.067 \pm 0.014$&$0.072 \pm 0.019$&$0.071 \pm 0.025$&$0.093 \pm 0.013$\\ 
 12 & $1.451 \pm 0.138$&$1.450 \pm 0.098$&$1.347 \pm 0.069$&$1.451 \pm 0.113$&$1.199 \pm 0.017$&$1.393 \pm 0.253$&$1.465 \pm 0.366$&$1.113 \pm 0.218$\\ 
 13 & $0.954 \pm 0.079$&$0.941 \pm 0.053$&$0.841 \pm 0.029$&$0.972 \pm 0.061$&$0.854 \pm 0.057$&$0.791 \pm 0.005$&$1.012 \pm 0.067$&$0.749 \pm 0.024$\\ 
 14 & $0.005 \pm 0.000$&$0.005 \pm 0.000$&$0.004 \pm 0.000$&$0.005 \pm 0.000$&$0.005 \pm 0.000$&$0.005 \pm 0.000$&$0.009 \pm 0.000$&$0.004 \pm 0.000$\\ 
 15 & $0.008 \pm 0.000$&$0.008 \pm 0.000$&$0.008 \pm 0.000$&$0.008 \pm 0.000$&$0.008 \pm 0.000$&$0.008 \pm 0.000$&$0.011 \pm 0.001$&$0.008 \pm 0.000$\\ 
 16 & $0.058 \pm 0.012$&$0.057 \pm 0.012$&$0.058 \pm 0.002$&$0.081 \pm 0.031$&$0.068 \pm 0.025$&$0.056 \pm 0.003$&$0.080 \pm 0.031$&$0.069 \pm 0.007$\\ 
 \bottomrule
    \end{tabular}}
    \caption{Per-client results on Graph-DT.}
    \label{tab:per_client2}
\end{table}

\begin{table}[th]
    \centering
    \resizebox{0.95\textwidth}{!}{
    \begin{tabular}{c c c c c c c c c}
    \toprule
Client ID & Isolated & FedAvg & FedAvg+FT & FedProx & FedBN & FedBN+FT & Ditto & FedMAML\\ 
\midrule
 1 & $0.870 \pm 0.008$ & $0.899 \pm 0.001$ & $0.899 \pm 0.001$ & $0.900 \pm 0.001$ & $0.899 \pm 0.001$ & $0.899 \pm 0.001$ & $0.898 \pm 0.001$ & $0.891 \pm 0.000$\\ 
 2 & $0.880 \pm 0.000$ & $0.889 \pm 0.000$ & $0.889 \pm 0.000$ & $0.888 \pm 0.001$ & $0.889 \pm 0.000$ & $0.889 \pm 0.000$ & $0.886 \pm 0.001$ & $0.889 \pm 0.003$\\
 3 & $0.675 \pm 0.005$ & $0.691 \pm 0.005$ & $0.696 \pm 0.005$ & $0.693 \pm 0.005$ & $0.695 \pm 0.005$ & $0.697 \pm 0.004$ & $0.720 \pm 0.000$ & $0.724 \pm 0.002$\\ 
 \bottomrule
    \end{tabular}}
    \caption{Per-client results on Text-DT.}
    \label{tab:per_client3}
\end{table}